\documentclass{article}

\PassOptionsToPackage{numbers, compress}{natbib}
\usepackage{listings}
\usepackage{xcolor}
\usepackage[main, final]{neurips_2026}
\usepackage{wrapfig}
\usepackage[utf8]{inputenc} 
\usepackage[T1]{fontenc}    
\usepackage{hyperref}       
\usepackage{url}            
\usepackage{booktabs}       
\usepackage{amsfonts}       
\usepackage{nicefrac}       
\usepackage{microtype}      
\usepackage{xcolor}         
\usepackage{graphicx}
\usepackage{subfig}
\usepackage{float} 
\usepackage[table]{xcolor}
\usepackage{multirow}
\usepackage{listings}
\usepackage{mdframed}
\newcommand{\thickunderline}[1]{\begingroup\sbox0{#1}\rlap{\rule[-0.6ex]{\wd0}{1pt}}\usebox0\endgroup}
\title{Exemplar2VQA: A Scalable Exemplar-Driven Visual Question Answering Generation Framework via Multi-Agent Coding}

\author{%
  Jiayu Ying$^{1}$, Qijian Tian$^{2}$, Ruijie Xu$^{1}$, Xinnan Zhu$^{1}$, Daoguo Dong$^{3}$, \\
  {\bfseries Jiachen Xu$^{4}$, and Xin Tan$^{1\dagger}$} \\
  $^{1}$East China Normal University \quad $^{2}$Shanghai Jiao Tong University \\
  $^{3}$Fudan University \quad $^{4}$Tencent \\
  \texttt{51295901118@stu.ecnu.edu.cn}, \texttt{xtan@cs.ecnu.edu.cn}, \texttt{tianqijian@sjtu.edu.cn} \\
  \texttt{\{ruijiex0, jcx97119, zxn.koala\}@gmail.com}, \texttt{dgdong@fudan.edu.cn}
}
\renewcommand{\thefootnote}{\ensuremath{\dagger}}
\renewcommand{\footnotemark}{\ensuremath{\dagger}}

\begin{document}

\maketitle
\footnotetext{Corresponding author.}
\renewcommand{\thefootnote}{\arabic{footnote}}
\begin{abstract}
    Advancing spatial intelligence in Multimodal Large Language Models (MLLMs) is bottlenecked by the scarcity of complex, scalable 3D question-answer (QA) data. While manual annotation is labor-intensive, directly utilizing LLMs to synthesize these QA pairs often fails due to their inherent deficiencies in spatial and geometric computation. We introduce \textbf{Exemplar2VQA}, a scalable exemplar-driven visual question answering generation framework that rapidly synthesizes large-scale spatial QA pairs in simulated environments via multi-agent coding. By equipping collaborative agents with a meticulously designed library of geometric utilities, Exemplar2VQA bypasses LLMs' spatial reasoning flaws through deterministic code execution. Crucially, the framework exhibits remarkable versatility: taking diverse static object-centric spatial query templates as exemplars, it seamlessly and autonomously scales them into massive, high-fidelity synthetic datasets. Fine-tuning Qwen2.5-VL (3B/7B) exclusively on Exemplar2VQA-generated synthetic indoor data yields significant performance improvements across various diverse benchmarks. Furthermore, its effectiveness is not limited to in-domain indoor datasets but also robustly extends to outdoor and mixed-scene benchmarks. These results establish Exemplar2VQA as a scalable and powerful paradigm for bridging the sim-to-real gap in Embodied AI. Our code is at \url{https://github.com/yingjiayu12/Exemplar2VQA}.
    
\end{abstract}

\section{Introduction}
    Spatial intelligence is crucial for MLLMs in Embodied AI, yet its advancement is severely bottlenecked by a scarcity of large-scale, high-quality 3D spatial QA datasets \cite{ScanQA}. Traditional manual annotation is prohibitively costly and unscalable. Conversely, using LLMs or VLMs to autonomously synthesize this frequently yields severe hallucinations \cite{SpatialVLM}. This stems from an inherent flaw: while proficient in language, LLMs lack deterministic geometric computational abilities, making them incapable of accurately grounding physical attributes like occlusion and allocentric viewpoints \cite{BLINK}. While multi-agent frameworks untangle this complexity by decomposing tasks \cite{AutoGen, MetaGPT}, merely distributing the workload fails to resolve LLMs' fundamental inability to perform precise spatial arithmetic.

    Consequently, synthesizing high-fidelity 3D spatial QA datasets at scale presents three formidable challenges: (1) The Data Acquisition Bottleneck: Traditional manual annotation is fundamentally unscalable and cost-prohibitive when dealing with complex, multi-view 3D geometries. (2) The Spatial Hallucination Dilemma: Directly prompting LLMs to deduce spatial attributes is highly prone to yielding logically flawed data \cite{11410321}. (3) The Rigidity of Existing Pipelines: Beyond the prevalent issue of hallucinations, current automated spatial generation methods are typically governed by fixed, hardcoded rules. They strictly lack the extensibility to dynamically adapt to user-defined abstract intents or flexibly generate customized QA pairs that align with diverse, specific structural formats.
    
    To address these formidable challenges, we propose Exemplar2VQA, designed to synthesize large-scale spatial QA pairs in simulated environments (Figure \ref{fig:pipeline_track1}, Figure \ref{fig:pipeline_track2}) without manual annotation. At the core of this synthesis process is the utilization of spatial QA exemplars as foundational blueprints. If a user wishes to construct a specific category of spatial questions, they merely need to provide a concrete QA pair as a exemplar. This exemplar functions provides a clear schema for the pipeline by explicitly defining the desired structural format, specific object parameters, and logical comparison targets. Our pipeline then autonomously scales this user-defined intent into a massive dataset. This seamless scaling from user-defined seeds effectively resolves the rigidity of existing automated pipelines. Guided by these templates, instead of forcing LLMs to generate linguistic spatial answers directly, our framework employs a collaborative multi-agent system that translates spatial reasoning tasks into executable Python scripts. Instead of forcing a monolithic model to suffer severe cognitive overload by simultaneously juggling semantic parsing, code generation, along with review and refinement, our collaborative compartmentalizes these distinct tasks into specialized agent roles. This decoupled architecture effectively dismantles the bottleneck of single-agent systems: it prevents the context contamination and compounding errors that inevitably occur when abstract linguistic intents are mixed with mathematical execution in a single prompt. By ensuring each agent maintains a highly focused reasoning window, this framework enables targeted, module-specific self-correction, thereby drastically improving the robustness and overall execution success rate \cite{DBLP:conf/aaai/YangLYLSL26}. Furthermore, by leveraging interactive environmental feedback from a physics-enabled simulator \cite{AI2-THOR, Habitat}, the agents can perform physics-constrained self-correction \cite{Code_as_Policies, Voyager, ViperGPT, DBLP:journals/tmlr/ChenM0C23}. This code-driven paradigm allows Exemplar2VQA to effectively bypass the inherent spatial reasoning flaws of conventional LLMs to derive reliable ground-truth answers \cite{Visual_Programming, luo2026pySpatial, DBLP:journals/corr/abs-2408-02193}.

    To empirically validate the efficacy of our generated data, we fine-tune open-weight multimodal large language models, Qwen2.5-VL (3B and 7B) \cite{qwen2.5-VL}, exclusively on the synthetic indoor data generated by Exemplar2VQA. The experimental results demonstrate substantial improvements not only on in-domain indoor benchmarks. Without seeing any outdoor training samples, the Exemplar2VQA-enhanced models successfully transfer their acquired spatial intelligence to strictly outdoor and comprehensive mixed-scene environments. Furthermore, our approach effectively outperforms baselines fine-tuned on human-annotated real-world datasets, underscoring a highly successful sim-to-real transfer \cite{DBLP:conf/iros/TobinFRSZA17, SynCLIP-AD}.
    
    In summary, the main contributions of this work are threefold:
    \begin{itemize}
    \item \textbf{An Exemplar-Driven Multi-Agent Spatial VQA Framework:} We introduce Exemplar2VQA, an end-to-end, fully automated VQA generation framework. By equipping collaborative agents with a meticulously designed library of geometric utilities and physics-constrained environmental feedback, Exemplar2VQA effectively resolves the inherent spatial computation deficiencies of LLMs through deterministic mathematical execution.
    \item \textbf{High Extensibility and Autonomous Scaling:} We demonstrate the exceptional versatility of Exemplar2VQA, showing that it can seamlessly adapt a broad spectrum of static object-centric spatial query exemplars from existing benchmarks and autonomously scale them into massive, high-quality synthetic datasets without manual intervention.
    \item \textbf{Robust Sim-to-Real and OOD Generalization:} Through extensive evaluations across diverse benchmarks, we validate that models fine-tuned solely on Exemplar2VQA's synthetic indoor data achieve extraordinary cross-domain spatial intelligence. This establishes a powerful, scalable, and cost-effective new paradigm for acquiring high-fidelity spatial data to advance Embodied AI.
    \end{itemize}

\section{Related Work}
    \textbf{Spatial Understanding in Vision-Language Models} has rapidly evolved from single-image reasoning \cite{DBLP:journals/tacl/0001EC23} to complex multi-view and cross-image scenarios \cite{BLINK, wang2024muirbench, DBLP:conf/eccv/TianYWXMT26}. To systematically diagnose spatial intelligence, recent benchmarks such as UniQA-3D \cite{DBLP:conf/3dim/ZuoKWJDG25}, ViewSpatial-Bench \cite{DBLP:journals/corr/abs-2505-21500}, and Multi-SpatialMLLM \cite{DBLP:journals/corr/abs-2505-17015} have shifted focus towards 3D and video-based environments. Notably, \citet{DBLP:conf/cvpr/YangYGH0X25} conducted comprehensive evaluations on VSI-Bench, revealing a stark reality: even advanced commercial MLLMs perform mediocrely on 3D spatial tasks. Existing spatial datasets face a critical dilemma: they either rely on fully human-curated annotations---such as MMSI-Bench proposed by \citet{DBLP:journals/corr/abs-2505-23764}, or depend on rigid, rule-based generation \cite{CLEVR}.
    
    \textbf{Automated Synthetic Data Generation} has emerged as a transformative paradigm to circumvent the prohibitive costs of manual annotation. Pioneering frameworks, such as LLaVA \citep{DBLP:conf/nips/LiuLWL23a} and ShareGPT4V \cite{ShareGPT4V}, have successfully leveraged powerful models like GPT-4 to autonomously synthesize large-scale multimodal instruction-tuning data. However, while this prompting-based synthesis excels in general semantic recognition, extending it to generate complex spatial QA pairs is notoriously problematic. When directly prompted to deduce precise 3D geometric relationships, MLLMs frequently suffer from severe spatial hallucinations \citep{DBLP:conf/icra/ChangPS25, DBLP:conf/emnlp/LiDZWZW23, Hallusionbench}. Consequently, directly prompting them to implicitly ``hallucinate'' physical relationships yields spatial data riddled with logical inconsistencies. Exemplar2VQA completely subverts this flawed paradigm.
    
    \textbf{Embodied Environments and Sim-to-Real Transfer} have become indispensable for scalable spatial learning in Embodied AI \cite{Sim2Real_Predictivity}. 3D physics-enabled simulators, such as AI2-THOR \citep{AI2-THOR} and Habitat \citep{Habitat}, offer an unparalleled testbed for generating high-fidelity training data. These platforms provide cost-effective, infinitely scalable \cite{ProcTHOR, Holodeck}, and fully controllable environments where perfect scene metadata can be effortlessly extracted to construct ground-truth spatial relationships \cite{3D-LLM}. Capitalizing on this pristine simulated data, Exemplar2VQA establishes a robust and deterministic foundation for spatial reasoning.

\section{Method} \label{Method}
In this section, we present Exemplar2VQA. Designed for the autonomous synthesis of high-fidelity spatial QA datasets, Exemplar2VQA utilizes concrete spatial query exemplars as guiding seeds.  We also detail the framework's overarching design, including the deterministic geometric utilities API it employs and the multi-agent collaborative pipeline that drives the dual-track synthesis process \cite{Toolformer, Gorilla, Code_as_Policies, PAL}.

\subsection{Problem Formulation and Exemplar2VQA Overview}
The fundamental objective of Exemplar2VQA is to autonomously synthesize large-scale, high-fidelity spatial QA datasets without relying on manual annotation or the hallucination-prone implicit reasoning of MLLMs. We formulate this synthesis process as a deterministic mapping. Formally, given a 3D simulated environment $\mathcal{S}$ and an abstract spatial query exemplar $\mathcal{E}$, Exemplar2VQA acts as an automated framework, denoted as $\Phi_{\mathrm{E2V}}$ for mathematical brevity, to generate a dataset $\mathcal{D} = \{(\mathcal{V}_i, Q_i, A_i)\}_{i=1}^N$. Here, $\mathcal{V}_i$ denotes a set of images captured according to the user-specified camera viewpoint requirements, $Q_i$ is the instantiated natural language question, and $A_i$ is the ground-truth answer. Thus, the pipeline can be abstracted as $\mathcal{D} = \Phi_{\mathrm{E2V}}(\mathcal{S}, \mathcal{E})$.

\begin{figure}
    \centering
    \includegraphics[width=\linewidth]{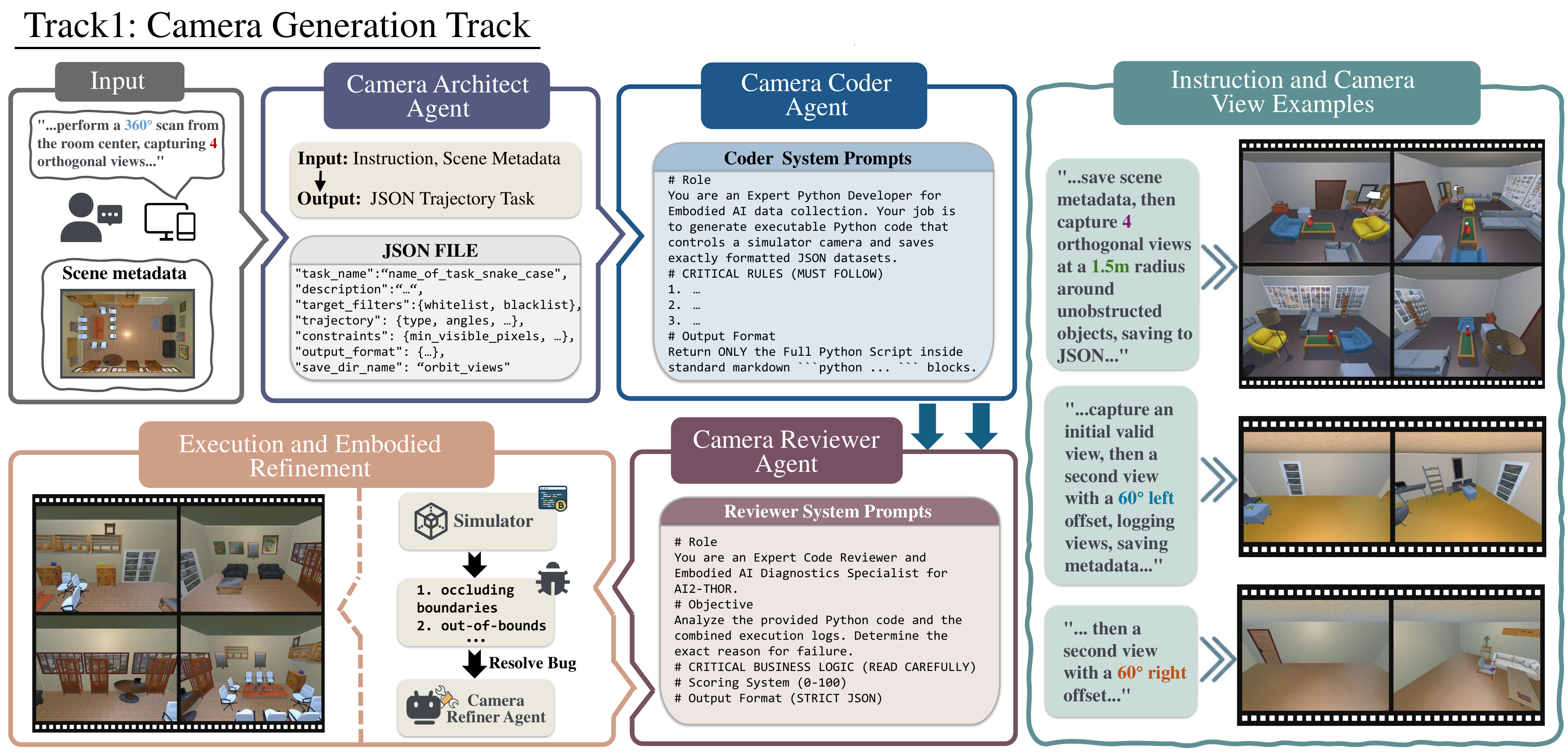}
    \caption{The camera generation track of the Exemplar2VQA. The Camera Generation Track: A four-agent pipeline (Architect, Coder, Reviewer, Refiner) interprets natural language instructions to autonomously plan trajectories and capture specific multi-view observations within the simulator.}
    \label{fig:pipeline_track1}
    \vspace{-10pt}
\end{figure}

\begin{figure}
    \centering
    \includegraphics[width=\linewidth]{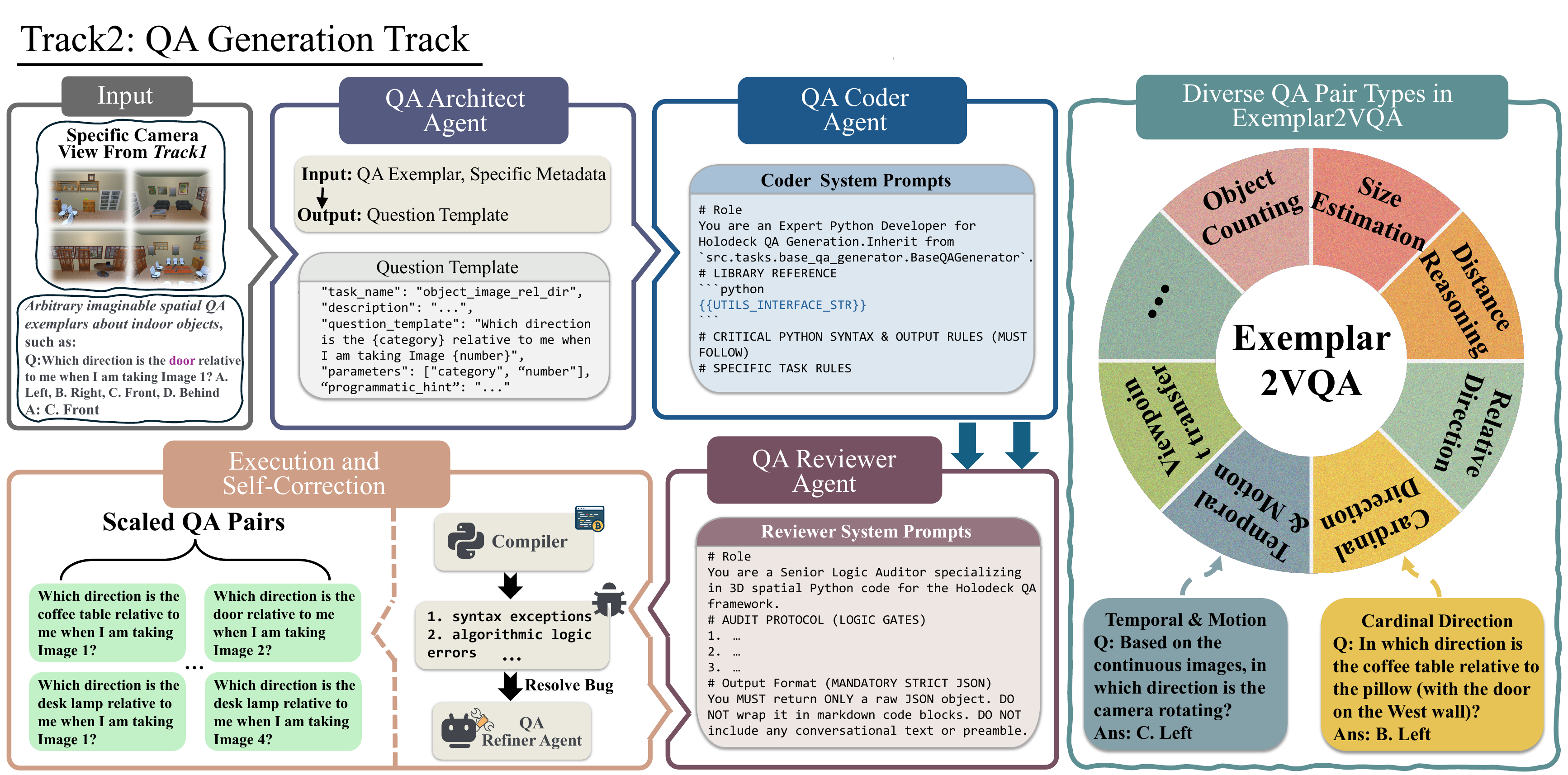}
    \caption{The QA generation track of the Exemplar2VQA. The QA Generation Track: Utilizing the captured scenes and scene metadata, an agent group adapts QA exemplars into scaled QA pairs.}
    \label{fig:pipeline_track2}
    \vspace{-12pt}
\end{figure}

To operationalize this mapping $\Phi_{\mathrm{E2V}}$, we introduce a logic-driven, dual-track multi-agent architecture, as illustrated in Figure \ref{fig:pipeline_track1}, \ref{fig:pipeline_track2}. Rather than forcing agents to implicitly guess spatial relationships, Exemplar2VQA conceptualizes them as code generators. The framework is bifurcated into two tracks.

\textbf{Camera Generation Track:} This track is responsible for navigating the environment $\mathcal{S}$ to capture valid and complex visual observations $\mathcal{V}_i$.

\textbf{QA Generation Track:} Utilizing the exact scene metadata (e.g., bounding boxes), this track programmatically derives $Q_i$ and mathematically computes $A_i$.

Crucially, by abstracting spatial reasoning into parameterized programmatic execution, Exemplar2VQA bypasses the constraints of rigid, rule-based generation. As demonstrated in Figure \ref{fig:pipeline_track2}, this architecture allows Exemplar2VQA to seamlessly adapt to a highly diverse spatial question taxonomy $\mathcal{E}$. From simple egocentric positioning to complex allocentric viewpoint reasoning and occlusion relationships, the framework autonomously scales these exemplars into massive, varied QA pairs, entirely eliminating the need for manual scripting.

    \subsection{Dual-Track Multi-Agent Collaborative Coding} \label{sec:multi_agent}

    To operationalize the deterministic mapping $\Phi_{\mathrm{E2V}}$, Exemplar2VQA orchestrates a logic-driven, dual-track multi-agent framework. Both tracks share a unified four-stage topology, but they operate within distinct execution environments to decouple embodied navigation from abstract spatial reasoning. All agents across both tracks are instantiated by a single foundational model: the \textbf{open-source} \texttt{Qwen3-Coder-30B-A3B-Instruct} \cite{qwen3technicalreport}. By leveraging a highly capable yet parameter-efficient (30B) open-source model rather than relying on massive, opaque proprietary LLMs, Exemplar2VQA ensures both high accessibility and full reproducibility. The distinct agent roles are strictly enforced not by different model weights, but through specialized system prompts, compartmentalized contextual memory, and track-specific tool access. This homogeneous design maintains consistent programmatic proficiency across the pipeline while preventing the context contamination that typically plagues monolithic agents.
    
    \textbf{Track 1: Camera Generation Track.}
    This track is responsible for autonomously navigating the simulator to capture valid and complex multi-view observations $\mathcal{V}_i$. To reliably bridge the semantic gap between abstract human intent and deterministic physical execution, we decompose the navigation pipeline into three distinct phases handled by specialized collaborative agents.

    \textbf{Stage 1: Semantic Parsing.} Given a natural language instruction $I_{\mathrm{NL}}$ (e.g., ``\textit{perform a 360$^\circ$ scan around the bed}''), the \textit{CameraArchitect} agent operates as a data extraction and planning module. It translates the abstract query into a structured, standardized JSON trajectory task $\mathcal{T}_{\mathrm{json}}$. This abstraction mathematically parametrizes the required embodied actions, explicitly defining target object filters $\mathcal{F}$, trajectory typologies, viewing angles $\theta$, and rigorous visual constraints $\mathcal{C}$ (e.g., minimum visible pixels). To handle the varying scale of indoor elements, the architect defines an adaptive radius parameter $r$ for the camera relative to a target object $o$, which can be formulated as:
    \begin{equation}
    r(o) = r_{\mathrm{base}} + \frac{1}{2} \max(D_o)
    \end{equation}
    where $r_{\mathrm{base}}$ is a predefined clearance scalar and $D_o \in \mathbb{R}^3$ represents the physical dimensions (length, width, height) of the target object's bounding box. Conversely, for tasks that do not target specific instances, this object-relative adaptive radius is bypassed.

    \textbf{Stage 2: Camera Code Generation.} Upon receiving the parameterized configuration $\mathcal{T}_{\mathrm{json}}$, the \textit{CameraCoder} agent acts as a geometric compiler, transforming the logical plan into an executable Python navigation script $z_{\mathrm{cam}}$. This script programmatically calculates the precise sequence of camera poses $P \in SE(3)$ required to capture the scene based on the designated trajectory typology. The generated code directly invokes the underlying simulator APIs to execute the embodied navigation and capture the required multi-view visual observations $\mathcal{V}_i$.
    
    \textbf{Stage 3: Execution and Embodied Refinement.} Crucially, this generated script $z_{\mathrm{cam}}$ interacts directly with the AI2-THOR physics engine. Due to the inherent unpredictability and geometric complexity of 3D environments, initial spatial scripts frequently trigger runtime failures, such as physical collisions with occluding boundaries or out-of-bounds rendering. When such embodied constraints are violated, the simulator returns a traceback error $e_{\mathrm{sim}}$. 

    To systematically resolve these physical conflicts, we introduce a decoupled diagnostic framework. The \textit{CameraReviewer} agent analyzes the execution trace and error state to diagnose the specific geometric or logical bug. Subsequently, the \textit{CameraRefiner} agent autoregressively patches the navigation code. This iterative refinement process can be mathematically formalized as:
    \begin{equation}
    z_{\mathrm{cam}}^{(k+1)} = \mathcal{R}_{\mathrm{cam}}\left(z_{\mathrm{cam}}^{(k)}, e_{\mathrm{sim}}^{(k)}\right)
    \end{equation}
    where $k$ denotes the refinement iteration index. This physics-constrained, closed-loop simulator feedback effectively eliminates invalid observations.

    \textbf{Track 2: QA Generation Track.}
    Operating downstream of Track 1, this track utilizes the precise scene metadata $\mathcal{M}$ associated with the captured observations $\mathcal{V}_i$ to mathematically derive instantiated questions $Q_i$ and ground-truth answers $A_i$.

    \textbf{Stage 1: Semantic Parsing.} Given a concrete, user-provided spatial query exemplar $\mathcal{E}$, the \textit{QA Architect} acts as a reverse-engineering module to disentangle the underlying spatial logic. It abstracts the specific textual exemplar into a generalized, parameterized schema $\Phi_{\mathcal{E}}$. Specifically, specific object names, numerical conditions, and image identifiers within $\mathcal{E}$ are replaced by a set of abstract placeholders $\mathbf{V} = \{v_{\mathrm{cat}}, v_{\mathrm{count}}, v_{\mathrm{view}}, \dots\}$ to formulate a highly scalable question template.

    \textbf{Stage 2: QA Code Generation.} Guided by the schema $\Phi_{\mathcal{E}}$, the \textit{QA Coder} synthesizes a Python script $z_{\mathrm{qa}}$. It invokes our meticulously engineered library of geometric utilities (detailed in Section \ref{sec:api}) to map the spatial reasoning tasks into explicit mathematical operations over the metadata $\mathcal{M}$.

    \textbf{Stage 3: Execution and Self-Correction.} During execution within the Python runtime environment, the initial script may encounter syntax exceptions, data-type mismatches, or algorithmic logic errors, triggering a runtime error $e_{\mathrm{py}}$. The \textit{QA Reviewer} and \textit{QA Refiner} systematically diagnose and debug these faults through an iterative autoregressive refinement loop:
    \begin{equation}
    z_{\mathrm{qa}}^{(k+1)} = \mathcal{R}_{\mathrm{qa}}\left(z_{\mathrm{qa}}^{(k)}, e_{\mathrm{py}}^{(k)}\right)
    \end{equation}
    Once a structurally and logically sound script $z_{\mathrm{qa}}^*$ is compiled, it systematically iterates over the permutations of the scene metadata $\mathcal{M}$ to compute and instantiate a massive, high-fidelity dataset of QA pairs $\mathcal{D}_{\mathrm{qa}} = \{(Q_i, A_i)\}_{i=1}^N$.

\subsection{Geometric API Orchestration}
\label{sec:api}
To bridge the gap between abstract spatial reasoning and physical reality, Exemplar2VQA equips the multi-agent system with a meticulously engineered library of deterministic geometric utilities. As fundamentally auto-regressive models, LLMs notoriously struggle with complex matrix transformations, and precise 3D geometric intersections, which are the primary catalysts for spatial hallucinations. Exemplar2VQA circumvents this architectural limitation by strictly offloading all rigorous geometric computations to a Python-based execution environment.

\definecolor{onedarkPurple}{HTML}{8931B9}
\definecolor{onedarkBlue}{HTML}{0D61AC}
\definecolor{onedarkGreen}{HTML}{3A8A24}  
\renewcommand{\lstlistingname}{Code}
\begin{wrapfigure}{r}{0.48\textwidth}
\vspace{-14pt}
\begin{lstlisting}[
    language=Python, 
    caption={\textbf{Selected} Exemplar2VQA \textbf{Geometric API}.}, 
    label={code:api}, 
    captionpos=b, 
    basicstyle=\ttfamily\tiny, 
    frame=single, 
    backgroundcolor=\color{gray!5}, 
    keywordstyle=\color{onedarkPurple}\bfseries, 
    commentstyle=\color{onedarkGreen}, 
    stringstyle=\color{onedarkGreen}, 
    emph={Exemplar2VQA_Utils, get_object_xz_points, calculate_planar_angles, is_spatial_relation_satisfied, calculate_projected_distance, is_angle_ambiguous, cal_3d_bbox_distance_between_categories}, 
    emphstyle=\color{onedarkBlue}\bfseries, 
    linewidth=\linewidth,
    breaklines=true,
    breakatwhitespace=false,
    keepspaces=true
]
class Exemplar2VQA_Utils:
  """geometric API."""

  def get_object_xz_points(obj_data, inc_corners=False):
      # Maps 3D 'obj_data' into a 2D [X, Z] coordinate footprint.

  def calculate_planar_angles(vec_ref, vec_targets):
      # Finds the relative angle of 'vec_targets' from 'vec_ref'.

  def is_spatial_relation_satisfied(dir, ref, tgt, th=0.15):
      # Validates if 'tgt' is strictly [dir] of 'ref'.

  def calculate_projected_distance(cam, tgt, angle):
      # Measures if moving at 'angle' gets 'cam' closer to 'tgt'.

  def is_angle_ambiguous(angles, threshold=10):
      # Rejects 'angles' near axes to prevent vague spatial queries.

  def cal_3d_bbox_distance_between_categories(A, B):
      # Finds the shortest 3D physical distance between groups 'A' and 'B'.
        
\end{lstlisting}
\vspace{-25pt}
\end{wrapfigure}

    As formalized in Code \ref{code:api}, we expose a structured hierarchy of spatial API signatures to the code-generation agents. Rather than treating these utilities as isolated tools, the \textit{QA Coder} synthesizes a programmatic execution plan, strictly enforcing geometric prerequisite chains. The LLM is abstracted away from continuous-space physics; its sole responsibility is topological routing and dependency orchestration. For instance, evaluating an allocentric directional relation mandates a strict order of operations: the pipeline must first invoke foundational APIs like \texttt{get\_object\_xz\_points} to project raw 3D scene metadata into 2D coordinate footprints. Only upon completing this prerequisite projection can the higher-level relational API, \texttt{is\_spatial\_relation\_satisfied}, perform deterministic interval intersections.
    
    Furthermore, the execution plan includes built-in geometric filters to maintain the high quality of the dataset. For instance, the system automatically routes calculated results through heuristic checks like \texttt{is\_angle\_ambiguous}. This step acts as a mandatory safety mechanism, actively discarding confusing or borderline spatial layouts before they can be generated into final QA pairs. By conceptualizing the LLM as a high-level coordinator that orchestrates these reliable mathematical functions, Exemplar2VQA significantly mitigates the risk of spatial hallucinations. 

\section{Experiments} \label{Experiments}

\subsection{Experimental Settings}

\textbf{Primary Evaluation Benchmarks.} We mainly evaluate our approach on MMSI-Bench \cite{DBLP:journals/corr/abs-2505-23764} and VSI-Bench \cite{DBLP:conf/cvpr/YangYGH0X25} to validate the direct performance improvements yielded by our synthesized data. Specifically, MMSI-Bench is a challenging VQA benchmark, comprising 1,000 meticulously crafted multiple-choice questions. VSI-Bench evaluates the visual-spatial intelligence of MLLMs from egocentric videos, featuring over 5,000 QA pairs derived from 288 real-world videos across diverse environments. Furthermore, comprehensive details regarding the adaptation of query exemplars from other benchmarks, along with their experimental data, are documented in Appendix \ref{sec:A.3}.

\begin{table}[htbp]
    \centering
\caption{\looseness=-1 \textbf{Performance comparison on the MMSI-Bench dataset}. The best results are shown in \textbf{bold}, and the second-best are \thickunderline{underlined}. Metrics marked with an asterisk (*) are evaluated in a strictly zero-shot setting. A superscript plus sign (\textbf{\textsuperscript{+}}) indicates a performance improvement over the base Qwen2.5-VL-7B model.}
\vspace{3pt}
\small
\setlength{\tabcolsep}{4pt} 
\resizebox{\textwidth}{!}{
\begin{tabular}{l cccccccccccc}
\toprule
\textbf{Method} & \textbf{Overall} & \textbf{Cam.-Cam.} & \textbf{Cam.-Obj.} & \textbf{Cam.-Reg.} & \textbf{Obj.-Obj.} & \textbf{Obj.-Reg.} & \textbf{Reg.-Reg.} & \textbf{Meas.} & \textbf{Cam.} & \textbf{MSR.} & \textbf{Appr.*} & \textbf{Obj.*} \\
\midrule

\rowcolor{blue!10}
\multicolumn{13}{l}{\textit{Baseline}} \\
InternVL3-38B \cite{InternVL3}                  & 26.3 & 21.5 & 23.3 & 25.3 & 20.2 & 35.3 & 33.3 & \textbf{39.1} & 16.2 & 25.8 & 21.2 & 31.6 \\

InternVL2.5-38B \cite{InternVL2.5}    & 27.9 & 18.3 & 22.1 & \thickunderline{34.9} & 22.3 & \textbf{38.8} & \thickunderline{35.8} & \thickunderline{37.5} & 14.9 & 25.3 & 25.8 & \thickunderline{38.2} \\

Qwen2.5-VL-32B \cite{qwen2.5-VL}     & 27.7 & 24.7 & 22.1 & 31.3 & \thickunderline{26.6} & 32.9 & 29.6 & 31.2 & 18.9 & 27.8 & 24.2 & 35.5 \\

InternVL3-8B \cite{InternVL3}       & 25.7 & 25.8 & 25.6 & 28.9 & \textbf{31.9} & 35.3 & \textbf{37.0} & 23.4 & 16.2 & 14.6 & 24.2 & 32.9 \\

DeepSeek-VL2 \cite{DeepSeek-VL2}       & 27.1 & 23.7 & \textbf{36.0} & 22.9 & \textbf{31.9} & 30.6 & 22.2 & 28.1 & \textbf{28.4} & \thickunderline{28.3} & 15.2 & 26.3 \\
\midrule

Qwen2.5-VL-7B (Base) \cite{qwen2.5-VL}   & 25.9 & 24.7 & 25.6 & 26.5 & 24.5 & 29.4 & 24.7 & 25.0 & 20.3 & 25.8 & 18.2 & \textbf{39.5} \\
Qwen2.5-VL-7B (SpaCE-10) \cite{SpaCE-10}                 & 27.0 & 30.1 & 26.7 & 31.3 & \thickunderline{26.6} & \thickunderline{37.7} & 29.6 & 20.3 & 14.9 & 24.2 & \thickunderline{27.3} & 29.0 \\
Qwen2.5-VL-7B (ViewSpatial) \cite{DBLP:journals/corr/abs-2505-21500}       & 26.2 & 20.4 & 25.6 & 32.5 & 25.5 & 31.8 & 29.6 & 29.7 & 13.5 & 24.8 & \textbf{30.3} & 27.6 \\
\rowcolor{gray!10}
\textbf{Exemplar2VQA (Ours, LoRA)} & \textbf{28.2}\rlap{\textbf{\textsuperscript{+}}} & \textbf{33.3}\rlap{\textbf{\textsuperscript{+}}} & 30.2\rlap{\textbf{\textsuperscript{+}}} & 33.7\rlap{\textbf{\textsuperscript{+}}} & 25.5\rlap{\textbf{\textsuperscript{+}}} & 23.5 & 29.6\rlap{\textbf{\textsuperscript{+}}} & 26.6\rlap{\textbf{\textsuperscript{+}}} & \thickunderline{23.0}\rlap{\textbf{\textsuperscript{+}}} & \textbf{29.3}\rlap{\textbf{\textsuperscript{+}}} & \thickunderline{27.3}\rlap{\textbf{\textsuperscript{+}}} & 25.0 \\  
\rowcolor{gray!10}
\textbf{Exemplar2VQA (Ours, Full SFT)} & \thickunderline{28.0}\rlap{\textbf{\textsuperscript{+}}} & \thickunderline{31.2}\rlap{\textbf{\textsuperscript{+}}} & \thickunderline{31.4}\rlap{\textbf{\textsuperscript{+}}} & \textbf{37.4}\rlap{\textbf{\textsuperscript{+}}} & 25.5\rlap{\textbf{\textsuperscript{+}}} & 24.7 & 32.1\rlap{\textbf{\textsuperscript{+}}} & 31.3\rlap{\textbf{\textsuperscript{+}}} & 18.9 & 25.8 & 21.2\rlap{\textbf{\textsuperscript{+}}} & 29.0 \\

\bottomrule
\end{tabular}
}
\label{tab:mmsibench}
\vspace{-15pt}

    \vspace{20pt}
    
    \caption{\looseness=-1 \textbf{Performance comparison on the VSI-Bench dataset}. 
\textbf{Left:} Quantitative results. The best results are shown in \textbf{bold}, and the second-best are \thickunderline{underlined}. Metrics marked with an asterisk (*) are evaluated in a strictly zero-shot setting. A superscript plus sign (\textbf{\textsuperscript{+}}) indicates a performance improvement over the base Qwen2.5-VL-3B model. 
\textbf{Right:} Bubble chart illustrating the relationship between zero-shot Route Plan performance and Average Performance.}
\vspace{3pt}
\label{tab:vsibench}

\begin{minipage}[c]{0.46\linewidth}
    \centering
    \small
    \setlength{\tabcolsep}{4pt}
    \resizebox{\linewidth}{!}{
    \begin{tabular}{l ccccc}
    \toprule
    \textbf{Method} & \textbf{Avg.} & \textbf{Rel. Dist.} & \textbf{Rel. Dir.} & \textbf{Route Plan*} & \textbf{Appr. Order} \\
    \midrule
    
    \rowcolor{blue!10}
    \multicolumn{6}{l}{\textit{Proprietary Models (API)}} \\
    GPT-4o \cite{GPT-4o}                 & 34.6 & 37.0 & 41.3 & 31.5 & 28.5 \\
    Gemini-1.5 Flash \cite{Gemini-1.5}       & 37.0 & 37.7 & 41.0 & 31.5 & 37.8 \\
    Gemini-1.5 Pro \cite{Gemini-1.5}         & \thickunderline{42.1} & \textbf{51.3} & 46.3 & \textbf{36.0} & 34.6 \\
    \midrule
    
    \rowcolor{blue!10}
    \multicolumn{6}{l}{\textit{Open-source Models}} \\
    LLaVA-Video-72B \cite{LLaVA-Video}        & 40.7 & 42.4 & 36.7 & 35.0 & \textbf{48.6} \\
    LLaVA-OneVision-72B \cite{LLaVA-OneVision}    & 39.9 & 42.5 & 39.9 & 32.5 & 44.6 \\
    InternVL2-40B \cite{InternVL2}          & 38.2 & \thickunderline{47.6} & 32.7 & 27.8 & 44.7 \\
    LLaVA-Video-7B \cite{LLaVA-Video}         & 37.6 & 43.5 & 42.4 & 34.0 & 30.6 \\
    InternVL2-8B \cite{InternVL2}           & 36.7 & 38.0 & 33.4 & 28.9 & \thickunderline{46.4} \\
    LLaVA-OneVision-7B \cite{LLaVA-OneVision}     & 32.9 & 42.5 & 35.2 & 29.4 & 24.4 \\
    \midrule
    
    Qwen2.5-VL-3B (Base) \cite{qwen2.5-VL}   & 35.3 & 34.7 & 42.6 & 28.9 & 35.0 \\
    Qwen2.5-VL-3B (MMSI-Bench) \cite{DBLP:journals/corr/abs-2505-23764}              & 35.2 & 36.8 & \thickunderline{47.2} & 29.4 & 27.2 \\
    Qwen2.5-VL-3B (SpaCE-10) \cite{SpaCE-10}                & 34.2 & 33.8 & 44.1 & 32.0 & 26.9 \\
    Qwen2.5-VL-3B (ViewSpatial) \cite{DBLP:journals/corr/abs-2505-21500}             & 37.8 & 33.8 & 46.7 & 33.0 & 37.7 \\
    \rowcolor{gray!10}
    \textbf{Exemplar2VQA (Ours)} & \textbf{42.9}\rlap{\textbf{\textsuperscript{+}}} & 43.8\rlap{\textbf{\textsuperscript{+}}} & \textbf{52.1}\rlap{\textbf{\textsuperscript{+}}} & \thickunderline{35.1}\rlap{\textbf{\textsuperscript{+}}} & 40.5\rlap{\textbf{\textsuperscript{+}}} \\
    
    \bottomrule
    \end{tabular}
    }
\end{minipage}\hfill
\begin{minipage}[c]{0.52\linewidth}
    \centering
    \includegraphics[width=\linewidth]{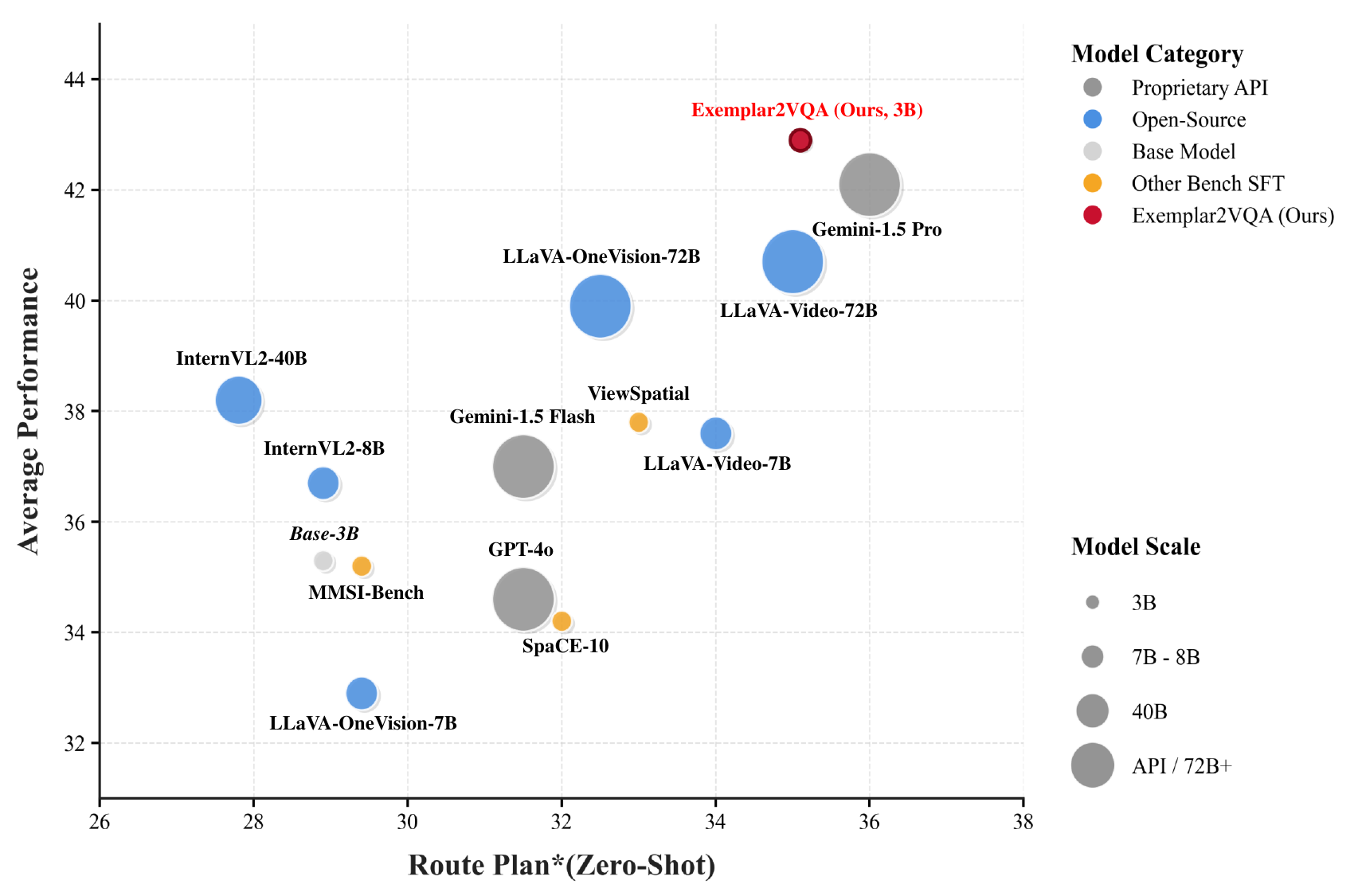}
\end{minipage}

    \vspace{-20pt}
\end{table}

\textbf{Zero-Shot Generalization Benchmarks.} To assess the cross-dataset transferability of the learned spatial priors, we employ SpaCE-10 \cite{SpaCE-10} and the challenging ViewSpatial-Bench \cite{DBLP:journals/corr/abs-2505-21500}. Notably, while SpaCE-10 tests generalization to unseen indoor topological queries, ViewSpatial-Bench introduces highly complex, mixed indoor and outdoor scenarios. Evaluations on an additional real-world life-scene benchmark, conducted under a similar zero-shot setup, are also provided in Appendix \ref{sec:A.3}.

\textbf{Baselines.} We compare the performance of Exemplar2VQA against four primary categories of baselines: (1) proprietary MLLMs, including GPT-4o \cite{GPT-4o}, Gemini-1.5 Pro \cite{Gemini-1.5}, and Gemini-1.5 Flash \cite{Gemini-1.5}; (2) open-weight MLLMs, such as LLaVA-Video-72B \cite{LLaVA-Video}, LLaVA-OneVision-72B \cite{LLaVA-OneVision}, InternVL2-40B \cite{InternVL2}, InternVL3-38B \cite{InternVL3}, InternVL2.5-38B \cite{InternVL2.5}, Qwen2.5-VL-32B \cite{qwen2.5-VL}, LLaVA-Video-7B \cite{LLaVA-Video}, LLaVA-OneVision-7B \cite{LLaVA-OneVision}, InternVL3-8B \cite{InternVL3}, InternVL2-8B \cite{InternVL2}, and DeepSeek-VL2 \cite{DeepSeek-VL2}; (3) the corresponding base models, Qwen2.5-VL-3B \cite{qwen2.5-VL} and Qwen2.5-VL-7B \cite{qwen2.5-VL}; and (4) spatial-specific baselines, where we fine-tune the base models on other existing human-annotated or rule-based spatial datasets (e.g., MMSI-Bench \cite{DBLP:journals/corr/abs-2505-23764}, SpaCE-10 \cite{SpaCE-10}, and ViewSpatial-Bench \cite{DBLP:journals/corr/abs-2505-21500}).

\textbf{Implementation Details.} For evaluations on MMSI-Bench \cite{DBLP:journals/corr/abs-2505-23764}, we synthesized approximately 6K QA pairs mirroring the target format and fine-tuned Qwen2.5-VL-7B \cite{qwen2.5-VL} using Low-Rank Adaptation (LoRA) \cite{LoRA} and Supervised Fine-Tuning (Full SFT) \cite{DBLP:conf/iclr/WeiBZGYLDDL22, DBLP:conf/nips/Ouyang0JAWMZASR22}. For VSI-Bench \cite{DBLP:conf/cvpr/YangYGH0X25}, we generated roughly 10K QA pairs and applied full-parameter SFT on Qwen2.5-VL-3B \cite{qwen2.5-VL}. To ensure balanced learning, we programmatically maintained a nearly uniform distribution across all spatial question types within these synthesized training datasets (see Appendix \ref{sec:A.4} for detailed). Furthermore, to guarantee the reliability of the training data, a subset of the synthesized QA pairs underwent manual spot-checking prior to fine-tuning. We deliberately capped the dataset generation at these scales (6K and 10K) to conserve computational resources, as preliminary training revealed that this relatively small amount of high-quality data is already sufficient to yield substantial performance gains. In practice, the Exemplar2VQA pipeline is structurally capable of generating an infinite number of QA pairs, bounded only by the diversity of the simulated 3D rooms. Exemplar2VQA can access additional rooms via ProcTHOR \cite{ProcTHOR} or Holodeck \cite{Holodeck} (see Appendix \ref{sec:A.2} for details). For all zero-shot evaluations on SpaCE-10 \cite{SpaCE-10} and ViewSpatial-Bench \cite{DBLP:journals/corr/abs-2505-21500}, we consistently deployed the Qwen2.5-VL-7B model fine-tuned solely on the MMSI-Bench synthetic data. It is imperative to note that throughout all experiments, we made absolutely no modifications to the underlying model architectures. To ensure a strictly fair comparison, all baseline models fine-tuned on other existing spatial datasets were trained using the exact same hyperparameters and optimization configurations as those applied to our Exemplar2VQA-generated data. We provide comprehensive implementation details in Appendix \ref{sec:A.1}.

\subsection{Performance on Primary Spatial Benchmarks}

\textbf{Results on MMSI-Bench.} 
Table \ref{tab:mmsibench} compares Exemplar2VQA against open-source MLLMs on the multi-image benchmark MMSI-Bench. Notably, tasks requiring abstract semantic shape projection (\textit{Appr.*}) or dynamic temporal tracking (\textit{Obj.*}) fall outside the static, coordinate-based generative scope of our current Exemplar2VQA pipeline. Therefore, we directly evaluated these two metrics under a zero-shot setting. Key observations include: (1) Ours achieves a new state-of-the-art overall score of 28.2\%, surpassing massive models like InternVL2.5-38B. Ours also outperforms the same 7B base model fine-tuned on other existing spatial datasets. (2) Ours substantially improves upon its base Qwen2.5-VL-7B model, with striking gains in complex multi-view reasoning (e.g., +8.6\% in Cam.-Cam.). (3) Even in the aforementioned zero-shot scenarios, Ours surprisingly boosts performance by 9.1\% on \textit{Appr.*}. (4) We observe a performance drop on \textit{Obj.*}. Fine-tuning exclusively on static spatial configurations biases the model toward static reasoning, which inevitably interferes with the base model's pre-trained temporal priors needed for dynamic tracking.  (5) To verify that the gain does not stem from the LoRA adaptation itself, we additionally fine-tune the same 7B model with full SFT on the identical training set, which yields a comparable overall score.

\textbf{Results on VSI-Bench.}
We further evaluate Exemplar2VQA on the egocentric video benchmark VSI-Bench (Table \ref{tab:vsibench}). Under our straightforward prompt-based SFT approach, directly training models for continuous numerical outputs yields sub-optimal results \cite{SpatialVLM}. Since our primary objective is to validate the effectiveness of Exemplar2VQA-generated data for spatial reasoning, we focus exclusively on the multiple-choice subset. Furthermore, tasks such as Route Planning (\textit{Route Plan*}), fall outside the static, point-to-point generative scope of our current pipeline and are thus evaluated in a zero-shot setting. From these evaluations, we observe that: (1) Our 3B model achieves an impressive 42.9\% overall, outperforming advanced proprietary models like GPT-4o and Gemini-1.5 Pro , as well as the 72B open-source LLaVA-Video. Crucially, Ours also decisively surpasses the same 3B base model fine-tuned on other existing spatial datasets. (2) Compared to its base Qwen2.5-VL-3B, Ours delivers a massive 7.6\% absolute boost in overall accuracy, with exceptional enhancements in specific sub-tasks like Relative Direction. (3) Even in the zero-shot \textit{Route Plan*} scenario, Ours yields a notable 6.2\% improvement over the base model. Moreover, the benefit of our synthetic data is not limited to a single backbone: fine-tuning InternVL2-2B and InternVL2-8B on the same training set yields consistent gains (see Table~\ref{tab:backbone_generalization} in Appendix), suggesting transferable supervision across model scales and architectures.

\begin{table}[t!]
    \centering
\caption{\looseness=-1 \textbf{Zero-shot performance comparison on the SpaCE-10 benchmark.} We evaluate the OOD generalization of our Exemplar2VQA-finetuned model against its base model and baselines fine-tuned on other spatial datasets. The best results are shown in \textbf{bold}. A superscript plus sign (\textbf{\textsuperscript{+}}) indicates a performance improvement over the base Qwen2.5-VL-7B model.}
\vspace{3pt}

\small
\setlength{\tabcolsep}{4pt} 
\resizebox{\textwidth}{!}{
\begin{tabular}{l ccccccccc}
\toprule
\textbf{Method} & \textbf{Overall} & \textbf{Entity} & \textbf{Scene} & \textbf{Size} & \textbf{Obj.-Obj.} & \textbf{Obj.-Scene} & \textbf{Entity} & \textbf{Function} & \textbf{Spatial} \\
& & \textbf{Quant.} & \textbf{Quant.} & \textbf{Assess.} & \textbf{Relation} & \textbf{Relation} & \textbf{Presence} & \textbf{Reasoning} & \textbf{Planning} \\
\midrule

Qwen2.5-VL-7B (Base) \cite{qwen2.5-VL} & 33.3 & 32.7 & \textbf{36.9} & 36.9 & 35.3 & 32.3 & 27.6 & 34.2 & 27.5 \\

Qwen2.5-VL-7B (MMSI-Bench) \cite{DBLP:journals/corr/abs-2505-23764} & 39.1 & \textbf{39.4} & 25.6 & 48.7 & 39.4 & 31.6 & 44.5 & 43.4 & 35.0 \\

Qwen2.5-VL-7B (ViewSpatial) \cite{DBLP:journals/corr/abs-2505-21500} & 40.6 & 36.1 & 14.3 & \textbf{58.4} & \textbf{47.7} & \textbf{35.6} & 41.1 & \textbf{51.5} & \textbf{42.5} \\

\rowcolor{gray!10}
\textbf{Exemplar2VQA (Ours)} & \textbf{42.2}\rlap{\textbf{\textsuperscript{+}}} & 38.4\rlap{\textbf{\textsuperscript{+}}} & 24.2 & 55.1\rlap{\textbf{\textsuperscript{+}}} & 45.5\rlap{\textbf{\textsuperscript{+}}} & 35.4\rlap{\textbf{\textsuperscript{+}}} & \textbf{45.9}\rlap{\textbf{\textsuperscript{+}}} & 50.3\rlap{\textbf{\textsuperscript{+}}} & 32.5\rlap{\textbf{\textsuperscript{+}}} \\

\bottomrule
\end{tabular}
}
\label{tab:space10}
    \vspace{4px}
    \centering
\caption{\looseness=-1 \textbf{Zero-shot performance comparison on the ViewSpatial-Bench dataset}. Quantitative results demonstrating the OOD generalization of our Exemplar2VQA-finetuned model against its base model and baselines fine-tuned on other spatial datasets. The best results are shown in \textbf{bold}. A superscript plus sign (\textbf{\textsuperscript{+}}) indicates a performance improvement over the base Qwen2.5-VL-7B model.}
\vspace{2pt}

\small
\setlength{\tabcolsep}{8pt}
\resizebox{0.9\linewidth}{!}{
\begin{tabular}{l c ccccc}
\toprule
\textbf{Method} & \textbf{Overall} & \textbf{Rel. Dir.} & \textbf{Obj. Ori.} & \textbf{Obj. Ori.} & \textbf{Rel. Dir.} & \textbf{Scene Sim.} \\
& & \textbf{(Cam)} & \textbf{(Cam)} & \textbf{(Per)} & \textbf{(Per)} & \textbf{(Per)} \\
\midrule

Qwen2.5-VL-7B (Base) \cite{qwen2.5-VL} & 36.9 & 46.6 & 29.7 & 37.1 & 35.0 & 28.8 \\

Qwen2.5-VL-7B (MMSI-Bench) \cite{DBLP:journals/corr/abs-2505-23764} & 36.5 & 45.3 & 29.7 & 41.4 & 35.6 & 24.7 \\

Qwen2.5-VL-7B (SpaCE-10) \cite{SpaCE-10} & 36.8 & 44.2 & 24.3 & 48.7 & 38.8 & 23.8 \\

\rowcolor{gray!10}
\textbf{Exemplar2VQA (Ours)} & \textbf{45.4}\rlap{\textbf{\textsuperscript{+}}} & \textbf{51.1}\rlap{\textbf{\textsuperscript{+}}} & \textbf{30.5}\rlap{\textbf{\textsuperscript{+}}} & \textbf{60.6}\rlap{\textbf{\textsuperscript{+}}} & \textbf{41.0}\rlap{\textbf{\textsuperscript{+}}} & \textbf{39.5}\rlap{\textbf{\textsuperscript{+}}} \\

\bottomrule
\end{tabular}
}
\label{tab:viewspatial}
    \vspace{-8pt}
\end{table}

\subsection{Performance on Zero-Shot Generalization Benchmarks}

\textbf{Results on SpaCE-10.}
To assess zero-shot OOD generalization, we evaluate Exemplar2VQA on the SpaCE-10 benchmark (Table \ref{tab:space10}). Key observations include: (1) Ours achieves an 8.9\% absolute overall improvement over the base Qwen2.5-VL-7B. Crucially, it also secures the highest overall zero-shot performance by outperforming the same base model fine-tuned on other spatial datasets. (2) Exceptional gains in physics- and geometry-heavy tasks, notably Size Assessment (+18.2\%) and Function Reasoning (+16.1\%). (3) The performance drop in Scene Quantification (\textit{Scene Quant.}) is an expected trade-off. This task requires abstract semantic grouping of "functional zones", causing slight negative transfer since Ours focuses strictly on instance-level bounding-box geometry. Nevertheless, overwhelming improvements in 7 out of 8 categories validate our approach.


\textbf{Results on ViewSpatial-Bench.}
Extending our zero-shot OOD evaluation to ViewSpatial-Bench (Table \ref{tab:viewspatial}), we observe: (1) Despite fine-tuning exclusively on synthetic indoor environments, ours delivers an 8.5\% overall boost on this diverse benchmark featuring complex mixed and outdoor scenes. Crucially, ours also significantly outperforms the same base model fine-tuned on other spatial datasets. (2) The most striking enhancement occurs in Object View Orientation from the human perspective (\textit{Obj. Ori. (Per)}), surging by 23.5\% absolute and easily eclipsing the other spatial baselines. (3) Consistent gains in other challenging person-centric tasks, such as Scene Simulation Relative Direction (+10.7\%) and Person-perspective Relative Direction (+6.0\%).

\begin{table}[t!]
    \centering
    \caption{\textbf{Ablation Study on Multi-Agent Configurations and Generator Capacity.} We evaluate the generation success rate (\%) given 100 seed examples under various agent topologies and generator backbones. Merged roles (indicated by '+') share the same context window, whereas separated roles (separated by ',') operate as distinct agents.}
    \vspace{4pt}
    \resizebox{\linewidth}{!}{
    \begin{tabular}{l cc}
    \toprule
    \textbf{Agent Configuration (Roles)} & \textbf{QA Generation (\%)} & \textbf{Camera Trajectory (\%)} \\
    \midrule
    \rowcolor{blue!10}
    \multicolumn{3}{l}{\textit{Agent Topology}} \\
    1 Agent (All roles merged) & 34.0 & 60.0 \\
    2 Agents (Architect, Coder+Reviewer+Refiner) & 72.0 & 74.0 \\
    3 Agents (Architect, Coder, Reviewer+Refiner) & 85.0 & 80.0 \\
    \rowcolor{gray!10}
    \textbf{4 Agents (Exemplar2VQA Full: Arch., Coder, Rev., Ref.)} & \textbf{92.0} & \textbf{84.0} \\
    \midrule
    \rowcolor{blue!10}
    \multicolumn{3}{l}{\textit{Generator Capacity (4 Agents)}} \\
    4 Agents (Qwen3.5-9B) & 48.0 & 56.0 \\
    4 Agents (Qwen3-Coder-30B) & \textbf{92.0} & \textbf{84.0} \\
    \bottomrule
    \end{tabular}
    }
    \label{tab:ablation_agents}
    

    \vspace{-8pt}
\end{table}
\begin{table}[t!]
    \centering
    \caption{\textbf{Comparison with Direct LLM Annotation.} We use Gemini3.5-Flash to directly annotate 10K synthetic training examples for the multiple-choice subset of VSI-Bench and fine-tune the same base model. Direct LLM annotation underperforms our code-based generation and even degrades the base model on average.}
    \vspace{4pt}
    \resizebox{\linewidth}{!}{
    \begin{tabular}{l ccccc}
    \toprule
    \textbf{Method} & \textbf{Rel. Dist} & \textbf{Rel. Dir} & \textbf{Route Plan} & \textbf{Appr. Order} & \textbf{Avg.} \\
    \midrule
    Qwen2.5-VL-3B (Base) & 34.7 & 42.6 & 28.9 & 35.0 & 35.3 \\
    Qwen2.5-VL-3B (LLM annotation) & 31.7 & 40.4 & 34.5 & 6.3 & 28.2 \\
    \rowcolor{gray!10}
    \textbf{Exemplar2VQA (Ours)} & \textbf{43.8} & \textbf{52.1} & \textbf{35.1} & \textbf{40.5} & \textbf{42.9} \\
    \bottomrule
    \end{tabular}
    }
    \label{tab:llm_annotation}
    \vspace{-8pt}
\end{table}

\subsection{Ablation Study on Multi-Agent Architectures and Generator Capacity}
\label{Ablation Study}

To validate the necessity of our decoupled multi-agent framework, we conduct an ablation study on the agent topology. We sample 100 seed examples and evaluate the generation success rate for both the QA Generation Track and the Camera Trajectory Track. Importantly, ``success'' strictly refers to the executed code producing mathematically and semantically correct final outputs. A comprehensive analysis of the error types is detailed in Section \ref{Error_Analysis}. As presented in Table \ref{tab:ablation_agents}, we progressively merge the specialized agent roles.

Key observations include: (1) \textbf{The failure of monolithic models.} When all roles are collapsed into a single agent, the accuracy plummets to 34.0\% for QA generation and 60.0\% for camera navigation. This empirically confirms that forcing a single LLM to simultaneously interpret semantics, write deterministic geometric code, and resolve physics-based simulator bugs leads to severe cognitive overload. (2) \textbf{The power of isolating semantic planning.} Splitting the pipeline into 2 agents yields massive absolute surges: +38.0\% in QA generation and +14.0\% in camera tracking. (3) \textbf{The necessity of decoupled debugging.} Progressively separating the bug diagnosis from the code patching process steadily pushes performance to its peak (92.0\% and 84.0\%).

Beyond topology, we further examine the impact of generator capacity. Since Qwen3-Coder-30B is already the smallest model in the Qwen3-Coder family, we adopt Qwen3.5-9B as a smaller generator alternative under the same 4-agent pipeline. This replacement drops QA generation from 92.0\% to 48.0\% and camera trajectory from 84.0\% to 56.0\%, suggesting that generator capacity is crucial for reliably synthesizing executable scripts and valid camera trajectories.

\subsection{Comparison with Direct LLM Annotation}
\label{sec:llm_annotation}

A natural alternative to our code-based generation is to directly employ a powerful VLM to annotate QA pairs. To investigate this, we use Gemini3.5-Flash to directly label 10K synthetic training examples for the multiple-choice subset of VSI-Bench, and fine-tune the same Qwen2.5-VL-3B base model on this annotated data under identical training configurations. As shown in Table \ref{tab:llm_annotation}, direct LLM annotation is insufficient for this setting: it underperforms Exemplar2VQA by a large margin and even degrades performance relative to the base model on average, with a particularly severe drop on Appr. Order. We attribute this to the fact that multi-view spatial QA demands not only locally plausible answer prediction, but also cross-view consistency, de-duplication, and stable context tracking, which cannot be guaranteed by direct LLM/VLM labeling.

\subsection{Remarks on Efficiency} \label{Remarks_on_Efficiency}
Operating across two GPUs, Exemplar2VQA maintains high efficiency for large-scale data synthesis. For given abstract spatial task, the code synthesis phase requires an average overhead: the Camera Trajectory Track averages only \textbf{3 minutes} to autonomously plan, code, and resolve physics-based bugs in AI2-THOR, while the QA Generation Track averages only \textbf{2 minutes} to synthesize the deterministic mathematical scripts. Once these reusable scripts are generated, the subsequent data execution scales linearly and rapidly with the number of simulated scenes without requiring further LLM intervention.  A comprehensive throughput analysis is provided in Appendix \ref{sec:Efficiency}.

\section{Conclusion}
In this work, we present Exemplar2VQA, a scalable exemplar-driven VQA generation framework that mitigates MLLMs' spatial reasoning limitations via deterministic mathematical execution. By leveraging geometric utilities and physics-constrained simulator feedback, Exemplar2VQA autonomously scales query exemplars into massive, high-fidelity spatial QA datasets. Experiments on MMSI-Bench and VSI-Bench demonstrate substantial performance gains over the base models, notably outperforming proprietary models like GPT-4o specifically on VSI-Bench. Furthermore, zero-shot evaluations on ViewSpatial-Bench and SpaCE-10 validate that spatial priors learned exclusively from synthetic indoor data effectively generalize to complex, unseen outdoor environments, offering a scalable approach to bridge the sim-to-real gap in Embodied AI.

\clearpage

{
\small
\bibliographystyle{unsrtnat}
\bibliography{ref}
}

\clearpage

\appendix

\section{Appendix}

\subsection{Implementation Details} \label{sec:A.1}

\subsubsection{Training Hyperparameters}

In this section, we detail the specific training configurations used to fine-tune our models on the Exemplar2VQA-generated synthetic datasets. Both models were trained utilizing the \texttt{swift} framework distributed. DeepSpeed ZeRO-2 optimization was employed in both setups to manage memory efficiency. Furthermore, to enhance instruction fine-tuning robustness, we applied NEFTune (Noise Embeddings for Fine-Tuning) with a noise alpha of 5 across all experiments. All inference evaluations were conducted on a single NVIDIA H200 GPU, while all fine-tuning procedures were distributed across four NVIDIA H200 GPUs. The dual-track Exemplar2VQA data generation pipelines were executed across two NVIDIA RTX A6000 GPUs. 

\textbf{Qwen2.5-VL-3B (VSI-Bench).} For the downstream VSI-Bench evaluation, we performed SFT on the Qwen2.5-VL-3B base model. The training was conducted over 3 epochs with a learning rate of 1e-5, utilizing a cosine learning rate scheduler with a warmup ratio of 0.05. We applied a weight decay of 0.01. The per-device train batch size was set to 4 with 2 gradient accumulation steps, resulting in a global effective batch size of 32 across the four GPUs.

\textbf{Qwen2.5-VL-7B (MMSI-Bench).} For the MMSI-Bench evaluation, we utilized LoRA to fine-tune the Qwen2.5-VL-7B base model using bfloat16 (bf16) precision. The LoRA rank ($r$) was set to 16 with a LoRA alpha of 32. Similar to the 3B model, training spanned 3 epochs, but with a higher learning rate of 1e-4. We employed a cosine learning rate scheduler, a warmup ratio of 0.05, and a weight decay of 0.01. The per-device train batch size was configured to 2 with 4 gradient accumulation steps, also yielding a global effective batch size of 32. To accommodate the specific visual demands of the multi-image dataset, we constrained the maximum image resolution to 1,003,520 pixels.

\subsubsection{Data Formatting and Training Prompts} \label{sec:data_prompts}

To ensure our models learn to strictly adhere to the diverse protocols of the target benchmarks during inference, we distinctively formatted our SFT training datasets based on the input modality requirements of each task.

\textbf{VSI-Bench.} To accurately align our synthetic training data with the evaluation format of VSI-Bench, which fundamentally assesses visual-spatial intelligence from egocentric videos, we converted the multi-view spatial images generated by Exemplar2VQA into continuous rotation videos. Consequently, the SFT dataset was formatted into a standard conversational structure containing a single video input sequence. The training instruction explicitly guides the model to output the deterministic spatial answer along with the text content, enclosed within precise XML-style tags.

\vspace{0.5em}
\noindent\textbf{Human Prompt Template (Training):}
\begin{mdframed}[
    topline=true,
    bottomline=true,
    rightline=true,
    leftline=false,
    linewidth=0.5pt,
    innerleftmargin=4pt,
    innerrightmargin=4pt,
    innertopmargin=4pt,
    innerbottommargin=4pt
]
{\small\ttfamily\raggedright
<video>

{[Task]} Your task is to analyze the spatial arrangement of objects in the scene by examining the provided video.

{[Answer Instruction]} You only need to provide *ONE* correct answer selecting from the options listed below. For example, if you think the correct answer is 'A. Above' from 'A. Above B. Under C. Front D. Behind', your response should **only** be '<answer>A. Above</answer>'.

{[Question]} \{question\_text\}
\par}
\end{mdframed}

\vspace{0.5em}
\noindent\textbf{Assistant Response:}
\begin{mdframed}[
    topline=true,
    bottomline=true,
    rightline=true,
    leftline=false,
    linewidth=0.5pt,
    innerleftmargin=4pt,
    innerrightmargin=4pt,
    innertopmargin=4pt,
    innerbottommargin=4pt
]
{\small\ttfamily\raggedright
<answer>\{answer\}</answer>
\par}
\end{mdframed}

\textbf{MMSI-Bench.} Conversely, MMSI-Bench evaluates spatial intelligence across a discrete sequence of multi-view images. To accommodate this, our SFT dataset formatting dynamically incorporates multiple \texttt{<image>} tokens corresponding to the exact number of input frames. The training instruction prompt is correspondingly adjusted to emphasize the analysis of spatial arrangements and motion across continuous discrete images, requiring the model to output solely the chosen answer letter to prevent hallucinated reasoning paths.

\vspace{0.5em}
\noindent\textbf{Human Prompt Template (Training):}
\begin{mdframed}[
    topline=true,
    bottomline=true,
    rightline=true,
    leftline=false,
    linewidth=0.5pt,
    innerleftmargin=4pt,
    innerrightmargin=4pt,
    innertopmargin=4pt,
    innerbottommargin=4pt
]
{\small\ttfamily\raggedright
<image> 
... (repeated N times based on sequence length)

{[Task]} Your task is to analyze the spatial arrangement and motion in the scene by examining the provided continuous images.

{[Answer Instruction]} You only need to provide *ONE* correct answer selecting from the options listed below. For example, if you think the correct answer is 'A. Above' from 'A. Above B. Under C. Front D. Behind', your response should **only** be '<answer>A. Above</answer>'.

{[Question]} \{question\_text\}
\par}
\end{mdframed}

\vspace{0.5em}
\noindent\textbf{Assistant Response:}
\begin{mdframed}[
    topline=true,
    bottomline=true,
    rightline=true,
    leftline=false,
    linewidth=0.5pt,
    innerleftmargin=4pt,
    innerrightmargin=4pt,
    innertopmargin=4pt,
    innerbottommargin=4pt
]
{\small\ttfamily\raggedright
<answer>\{answer\}</answer>
\par}
\end{mdframed}
\vspace{0.5em}

This explicit, modality-aware formatting strategy effectively guides the models during the fine-tuning phase to bypass verbose implicit reasoning and predictably output answers in a deterministic, automated-evaluation-friendly structure.

\subsubsection{Evaluation Prompts} \label{sec:eval_prompts}

During the downstream evaluation phase, we apply rigorous prompt templates to ensure the models are evaluated consistently across different benchmarks. These prompts strictly mirror the output formatting constraints established during the training phase, adapting only the modality tokens (video vs. images) to suit the respective benchmark.

\vspace{0.5em}
\noindent\textbf{Evaluation Prompt Template (VSI-Bench):}
\begin{mdframed}[
    topline=true,
    bottomline=true,
    rightline=true,
    leftline=false,
    linewidth=0.5pt,
    innerleftmargin=4pt,
    innerrightmargin=4pt,
    innertopmargin=4pt,
    innerbottommargin=4pt
]
{\small\ttfamily\raggedright
<video>

{[Task]} Your task is to analyze the spatial arrangement of objects in the scene by examining the provided video.

{[Answer Instruction]} You only need to provide *ONE* correct answer selecting from the options listed below. For example, if you think the correct answer is 'A. Above' from 'A. Above B. Under C. Front D. Behind', your response should **only** be '<answer>A. Above</answer>'.

{[Question]} \{full\_question\}
\par}
\end{mdframed}

\vspace{0.5em}
\noindent\textbf{Evaluation Prompt Template (MMSI-Bench):}
\begin{mdframed}[
    topline=true,
    bottomline=true,
    rightline=true,
    leftline=false,
    linewidth=0.5pt,
    innerleftmargin=4pt,
    innerrightmargin=4pt,
    innertopmargin=4pt,
    innerbottommargin=4pt
]
{\small\ttfamily\raggedright
<image> 
... (repeated N times based on sequence length)

{[Task]} Your task is to analyze the spatial arrangement of objects in the scene by examining the provided images.

{[Answer Instruction]} You only need to provide *ONE* correct answer selecting from the options listed below. For example, if you think the correct answer is 'A. Above' from 'A. Above B. Under C. Front D. Behind', your response should **only** be '<answer>A. Above</answer>'.

{[Question]} \{full\_question\}
\par}
\end{mdframed}

\subsection{Infinite Generation via ProcTHOR and Holodeck} \label{sec:A.2}

While our experiments deliberately capped the generated training datasets at 6K and 10K pairs to conserve computational resources, the Exemplar2VQA framework is fundamentally designed as a highly scalable engine. Because Exemplar2VQA relies entirely on deterministic programmatic execution interacting with simulator scene metadata, its generation capacity is bottlenecked solely by the diversity and quantity of available 3D simulated environments. To overcome the limitations of a fixed set of manually crafted scenes, Exemplar2VQA can seamlessly integrate with advanced environment generation platforms such as ProcTHOR \cite{ProcTHOR} and Holodeck \cite{Holodeck}, effectively unlocking an "infinite generation" paradigm.

\textbf{Scaling with ProcTHOR.} ProcTHOR serves as a massive repository of procedurally generated, ready-to-use 3D environments. It provides access to tens of thousands of diverse, fully interactive houses and floorplans. Exemplar2VQA can autonomously iterate through this extensive database, continuously extracting fresh scene metadata (e.g., novel object configurations, varying room layouts, and diverse occlusions) without requiring any manual curation. This ensures that the generated spatial question-answer pairs maintain high domain diversity and rarely suffer from geometric repetition.

\textbf{Customized Generation via Holodeck.} For targeted domain adaptation or handling highly specific spatial scenarios, Exemplar2VQA leverages Holodeck for dynamic, language-guided environment generation. Holodeck allows the system to synthesize entirely new rooms within the AI2-THOR simulator simply by providing a natural language room definition or prompt (e.g., "A co-working space featuring a large communal desk and multiple rolling chairs arranged in a grid pattern at the room’s center." or "A media room featuring a large recliner and a low TV stand arranged in a linear orientation at the room’s center."). Once the prompt is processed, Holodeck autonomously populates the layout and outputs the precise scene metadata.

By utilizing ProcTHOR for broad, large-scale data expansion and Holodeck for customized, prompt-driven scene creation, Exemplar2VQA establishes a virtually unlimited, continuously expanding sandbox. This synergy guarantees an endless supply of high-fidelity spatial configurations.

\subsection{Extended Experiments: Template Adaptation and QA Visualizations} \label{sec:A.3}

In this section, we present extended experimental results to further demonstrate the exceptional versatility and OOD generalization of the Exemplar2VQA framework. As discussed in the main text, Exemplar2VQA is not confined to a predefined set of tasks; rather, it can seamlessly adapt static object-centric spatial query templates from diverse downstream benchmarks. 

\subsubsection{Evaluation Results on OSR-Bench} 

\textbf{Omni-Spatial-Reasoning (OSR) Benchmarks \cite{DBLP:journals/corr/abs-2505-11907}.} To further investigate spatial intelligence in $360^{\circ}$ panoramic environments, we extend our evaluation to OSR-Bench.  OSR-Bench comprises over 153,000 diverse Question-Answer (QA) pairs. 

\textbf{Implementation Details.} For the extended evaluation on OSR-Bench, we synthesized approximately 6K QA pairs utilizing the Exemplar2VQA framework, meticulously adhering to the diverse spatial query templates established by the benchmark (including object counting, relative distance, and relative direction, as visualized in Figure \ref{fig:appendix_example}). We then fine-tuned the Qwen2.5-VL-7B model exclusively on this newly synthesized dataset. Consistent with our approach for MMSI-Bench, the fine-tuning was performed using Low-Rank Adaptation (LoRA) and distributed across four NVIDIA H200 GPUs. The specific training hyperparameters were kept strictly identical to those used for the MMSI-Bench evaluations to ensure comparative consistency. Comprehensive details regarding these training hyperparameters are provided in Appendix \ref{sec:A.1}.

\setcounter{figure}{0}
\renewcommand{\thefigure}{A\arabic{figure}}
\begin{figure}
    \centering
    \includegraphics[width=\linewidth]{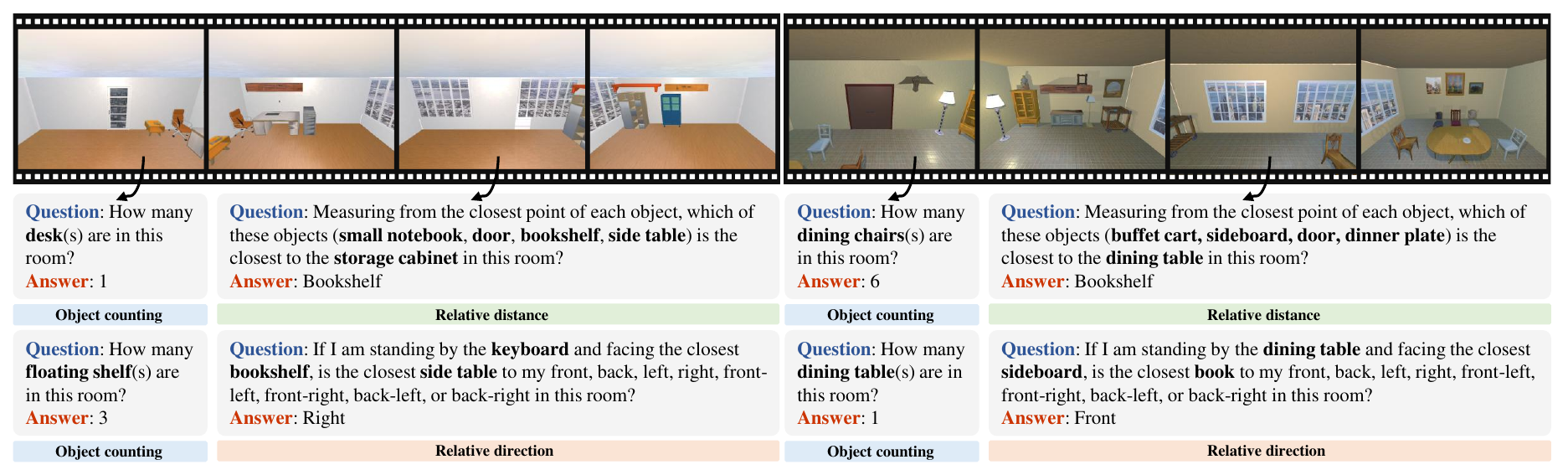}
    \caption{Visual examples of diversified spatial QA pairs and corresponding 3D scenes programmatically generated by the Exemplar2VQA framework. This visualization demonstrates the framework's versatility in adapting and scaling diverse query templates sourced from the \textbf{OSR-Bench} for \textbf{Object Counting}, \textbf{Relative Distance}, and \textbf{Relative Direction} across different generated environments to create massive and high-fidelity spatial reasoning datasets.}
    \label{fig:appendix_example}
    \vspace{-6pt}
\end{figure}

\setcounter{table}{0}
\renewcommand{\thetable}{A\arabic{table}}
\begin{table}[t!]
    \centering
\caption{\looseness=-1 \textbf{Performance comparison on the OSR-Bench dataset}. The best results are shown in \textbf{bold}, and the second-best are \thickunderline{underlined}. A superscript plus sign (\textbf{\textsuperscript{+}}) indicates a performance improvement over the base Qwen2.5-VL-7B model.}
\label{tab:osr_bench}
\vspace{3pt}
\small
\setlength{\tabcolsep}{4pt} 
\resizebox{\textwidth}{!}{
\begin{tabular}{l cccc}
\toprule
\textbf{Method} & \textbf{Avg.} & \textbf{Object Count} & \textbf{Relative Distance} & \textbf{Relative Direction} \\
\midrule

\rowcolor{blue!10}
\multicolumn{5}{l}{\textit{Baseline}} \\
Qwen2.5-VL-72B \cite{qwen2.5-VL}   & \thickunderline{33.5} & 49.8 & \thickunderline{32.5} & \textbf{18.1} \\

LLaVA-v1.5-13B \cite{DBLP:conf/cvpr/LiuLLL24}    & 31.6 & 55.3 & 22.6 & \thickunderline{16.9} \\

DeepSeek-VL2 \cite{DeepSeek-VL2}   & 25.5 & \thickunderline{57.9} & 5.5 & 13.1 \\
\midrule

Qwen2.5-VL-7B (Base) \cite{qwen2.5-VL} & 28.3 & 46.3 & 31.1 & 7.4 \\
\rowcolor{gray!10}
\textbf{Exemplar2VQA (Ours)} & \textbf{36.5}\rlap{\textbf{\textsuperscript{+}}} & \textbf{59.0}\rlap{\textbf{\textsuperscript{+}}} & \textbf{41.1}\rlap{\textbf{\textsuperscript{+}}} & 9.3\rlap{\textbf{\textsuperscript{+}}} \\  

\bottomrule
\end{tabular}
}
\label{tab:new_spatial_results}

    \vspace{-5pt}
\end{table}

\textbf{Results on OSR-Bench.} Table \ref{tab:osr_bench} compares Exemplar2VQA against powerful baseline MLLMs on the OSR-Bench dataset. Crucially, the training data synthesized by Exemplar2VQA consisted merely of four discrete, sparse surround-view images per scene, rather than the continuous equirectangular $360^{\circ}$ panoramas natively evaluated in OSR-Bench. Key observations include: (1) Despite this distinct visual domain gap, Exemplar2VQA (Ours, LoRA-7B) achieves an impressive overall average of 36.5\%, successfully outperforming significantly larger baseline models such as Qwen2.5-VL-72B (33.5\%) and LLaVA-v1.5-13B (31.6\%). Notably, it establishes top performance in both Object Count (59.0\%) and Relative Distance (41.1\%). (2) Exemplar2VQA delivers a massive 8.2\% absolute boost in overall accuracy compared to its base Qwen2.5-VL-7B model (from 28.3\% to 36.5\%), with consistent enhancements across all sub-tasks, including a striking +12.7\% increase in Object Count and +10.0\% in Relative Distance. 

While the Exemplar2VQA rendering pipeline is structurally capable of stitching and projecting continuous $360^{\circ}$ panoramic images for training—which would naturally eliminate this modality gap and likely drive performance even higher—we deliberately omitted this step. Because our primary objective is to validate the fundamental utility and geometric accuracy of the Exemplar2VQA pipeline's automated QA generation, rather than merely engineering a state-of-the-art model tailored to a specific image format, the current substantial performance gains under this challenging zero-shot view-transfer setting are already sufficient to prove our premise.

\begin{table}[t!]
    \centering
\caption{\looseness=-1 \textbf{Zero-shot performance comparison on the Spatial-Obj dataset}. Quantitative results demonstrating the spatial reasoning capabilities of our Exemplar2VQA-finetuned model on one-object and two-object scenarios. The best results are shown in \textbf{bold}. A superscript plus sign (\textbf{\textsuperscript{+}}) indicates a performance improvement over the base Qwen2.5-VL-7B model.}
\vspace{2pt}

\small
\setlength{\tabcolsep}{12pt}
\resizebox{0.7\linewidth}{!}{
\begin{tabular}{l ccc}
\toprule
\textbf{Method} & \textbf{1\_obj} & \textbf{2\_obj} & \textbf{Overall} \\
\midrule

Qwen2.5-VL-7B (Base) \cite{qwen2.5-VL} & 83.86 & 62.90 & 69.71 \\

\rowcolor{gray!10}
\rule{0pt}{1.1em}\textbf{Exemplar2VQA (Ours)} & \textbf{84.46}\rlap{\textbf{\textsuperscript{+}}} & \textbf{63.85}\rlap{\textbf{\textsuperscript{+}}} & \textbf{70.50}\rlap{\textbf{\textsuperscript{+}}} \\

\bottomrule
\end{tabular}
}
\label{tab:spatial_obj}
    \vspace{-5pt}
\end{table}

\renewcommand{\thetable}{A\arabic{table}}
\begin{table}[t!]
    \centering
\caption{\textbf{Generalization across backbone architectures.} We fine-tune InternVL2-2B and InternVL2-8B on the same synthetic dataset generated by Exemplar2VQA and evaluate on VSI-Bench. The best results are shown in \textbf{bold}, and the second-best are \thickunderline{underlined}. A superscript plus sign (\textbf{\textsuperscript{+}}) indicates a performance improvement over the corresponding base model.}
\vspace{4pt}
\setlength{\tabcolsep}{4pt}
\resizebox{0.86\linewidth}{!}{
\begin{tabular}{l ccccc}
\toprule
\textbf{Method} & \textbf{Avg.} & \textbf{Rel. Dist.} & \textbf{Rel. Dir.} & \textbf{Route Plan*} & \textbf{Appr. Order} \\
\midrule
GPT-4o \cite{GPT-4o}               & 34.6 & 37.0 & 41.3 & 31.5 & 28.5 \\
Gemini-1.5 Pro \cite{Gemini-1.5}   & \thickunderline{42.1} & \textbf{51.3} & 46.3 & \thickunderline{36.0} & 34.6 \\
InternVL2-2B \cite{InternVL2}      & 28.2 & 32.1 & 44.1 & 30.4 & 6.3 \\
InternVL2-8B \cite{InternVL2}      & 36.7 & 38.0 & 33.4 & 28.9 & \thickunderline{46.4} \\
\midrule

\rowcolor{gray!10}
\textbf{Exemplar2VQA (Ours, InternVL2-2B)} & 35.6\rlap{\textbf{\textsuperscript{+}}} & 33.6\rlap{\textbf{\textsuperscript{+}}} & \textbf{47.6}\rlap{\textbf{\textsuperscript{+}}} & \thickunderline{36.0}\rlap{\textbf{\textsuperscript{+}}} & 25.2\rlap{\textbf{\textsuperscript{+}}} \\
\rowcolor{gray!10}
\textbf{Exemplar2VQA (Ours, InternVL2-8B)} & \textbf{46.1}\rlap{\textbf{\textsuperscript{+}}} & \thickunderline{50.4}\rlap{\textbf{\textsuperscript{+}}} & \thickunderline{47.2}\rlap{\textbf{\textsuperscript{+}}} & \textbf{36.1}\rlap{\textbf{\textsuperscript{+}}} & \textbf{50.6}\rlap{\textbf{\textsuperscript{+}}} \\
\bottomrule
\end{tabular}
}
\label{tab:backbone_generalization}

    \vspace{-5pt}
\end{table}

\subsubsection{Evaluation Results on Spatial-MM Datasets} 

\textbf{Spatial-MM Datasets \cite{DBLP:conf/emnlp/ShiriGF0HL24}.} To evaluate spatial reasoning capabilities in everyday environments, we utilize Spatial-MM, an \textbf{additional real-world life-scene} datasets. For our experiments, we focus exclusively on its Spatial-Obj subset. Spatial-Obj comprises 2,000 multiple-choice questions that assess spatial relationships involving one or two objects within an image. The dataset categorizes these spatial configurations into distinct visual patterns, including object localization, orientation and direction, viewpoints, and positional and relational context. Furthermore, the questions are structured to evaluate spatial relationships from both the standard camera perspective and the human perspective within the image.

\textbf{Implementation Details.} For the evaluation on the Spatial-MM benchmark, we adopted a strictly zero-shot setting, consistent with our methodology for the SpaCE-10 and ViewSpatial-Bench evaluations. Specifically, we deployed the Qwen2.5-VL-7B model that was fine-tuned exclusively on the synthetic data generated for MMSI-Bench. No additional, dataset-specific training or parameter updates were performed using the Spatial-MM data. This rigorous zero-shot setup allows us to genuinely assess the cross-dataset transferability of the spatial priors learned from the Exemplar2VQA pipeline. Comprehensive implementation details regarding the acquisition of this deployed model, including the specific training hyperparameters, are detailed in Appendix \ref{sec:A.1}.

\textbf{Results on Spatial-Obj Subset.} Table \ref{tab:spatial_obj} details the zero-shot performance of Exemplar2VQA on the Spatial-Obj datasets. Crucially, while the Exemplar2VQA model was fine-tuned exclusively on synthetic indoor environments, Spatial-Obj evaluates spatial reasoning across diverse, real-world life scenes. Key observations include: (1) Despite this challenging sim-to-real domain gap, Exemplar2VQA (Ours, LoRA-7B) consistently outperforms the Qwen2.5-VL-7B base model across all evaluated configurations. (2) Exemplar2VQA achieves steady improvements in both single-object (\textit{1\_obj}, +0.60\% absolute) and two-object (\textit{2\_obj}, +0.95\% absolute) spatial scenarios, yielding an overall accuracy boost to 70.50\%.

\begin{figure}
    \centering
    \includegraphics[width=\linewidth]{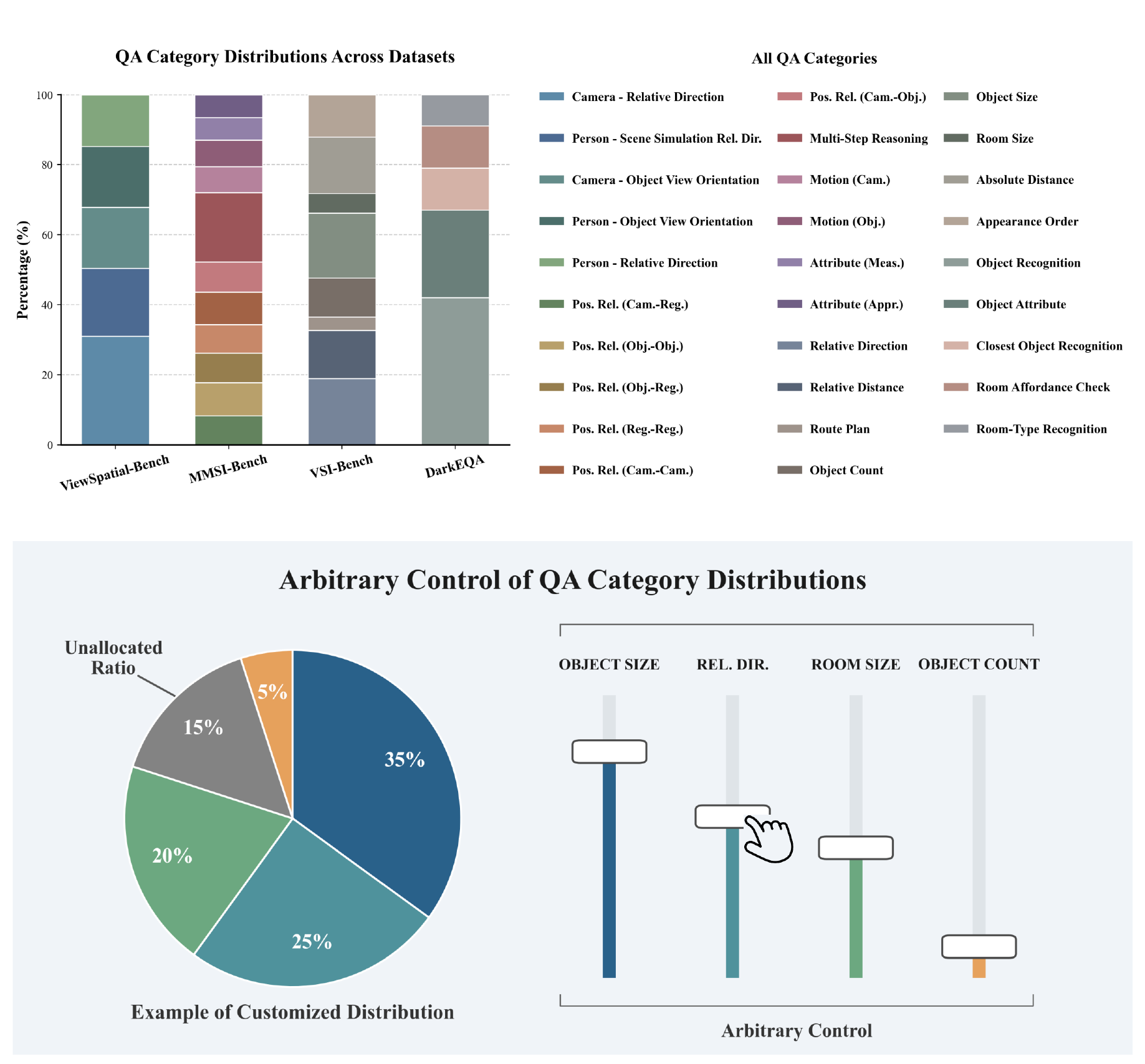}
    \caption{\textbf{Customizable QA category generation versus static benchmark distributions.} \textbf{(Top)} Existing spatial reasoning datasets (e.g., ViewSpatial-Bench, MMSI-Bench) exhibit fixed and inherently imbalanced QA category distributions. \textbf{(Bottom)} In contrast, our proposed pipeline enables arbitrary, programmatic control over the generation ratios. Driven by a theoretically infinite generation capacity—bounded only by the diversity of available 3D scenes—our framework allows users to dynamically configure the distribution of specific spatial queries. This paradigm effectively overcomes the static distribution biases prevalent in traditional human-annotated benchmarks.}
    \label{fig:compare}
    \vspace{-10pt}
\end{figure}

While the absolute numerical gains may appear modest compared to our in-domain evaluations, they are highly meaningful within this strict zero-shot context. The base model already exhibits a remarkably high performance baseline on this specific dataset (approaching 84\% on \textit{1\_obj}, indicating near-saturation). The ability of Exemplar2VQA to squeeze out further improvements on unseen, complex real-world photographs—relying strictly on geometric priors learned from synthetic multi-view data—confirms that our automated generation framework injects robust, universally transferable spatial intelligence rather than simply overfitting to the synthetic domain.

\subsubsection{Stability Across Random Seeds}
\label{app:mmsi_seeds}

To examine the stability of our MMSI-Bench results, we rerun the fine-tuning experiment with three different random seeds, using the same training data each time.
As shown in Table~\ref{tab:mmsi_seeds}, the repeated runs suggest that the reported gains are stable across seeds, although the magnitude of improvement varies across sub-categories.

\renewcommand{\thetable}{A\arabic{table}}
\begin{table}[t!]
    \centering
\caption{\textbf{Stability of MMSI-Bench results across random seeds.} We rerun the MMSI-Bench fine-tuning experiment with three different random seeds using the same training data each time, and report the mean with the standard deviation as a superscript. Metrics marked with an asterisk (*) are evaluated in a strictly zero-shot setting.}
\vspace{4pt}
\setlength{\tabcolsep}{4pt}
\resizebox{0.98\linewidth}{!}{
\begin{tabular}{l cccccccccccc}
\toprule
\textbf{Method} & \textbf{Overall} & \textbf{Cam.-Cam.} & \textbf{Cam.-Obj.} & \textbf{Cam.-Reg.} & \textbf{Obj.-Obj.} & \textbf{Obj.-Reg.} & \textbf{Reg.-Reg.} & \textbf{Meas.} & \textbf{Cam.} & \textbf{MSR.} & \textbf{Appr.*} & \textbf{Obj.*} \\
\midrule
\rowcolor{gray!10}
\textbf{Exemplar2VQA} & \textbf{29.33}\textsuperscript{\tiny$\pm$0.99} & 31.54\textsuperscript{\tiny$\pm$2.24} & 33.72\textsuperscript{\tiny$\pm$1.16} & 37.35\textsuperscript{\tiny$\pm$4.34} & 26.96\textsuperscript{\tiny$\pm$2.66} & 30.98\textsuperscript{\tiny$\pm$2.96} & 30.04\textsuperscript{\tiny$\pm$1.42} & 30.73\textsuperscript{\tiny$\pm$4.78} & 20.27\textsuperscript{\tiny$\pm$2.34} & 28.96\textsuperscript{\tiny$\pm$1.05} & 24.24\textsuperscript{\tiny$\pm$0.00} & 25.00\textsuperscript{\tiny$\pm$1.32} \\
\bottomrule
\end{tabular}
}
\label{tab:mmsi_seeds}

    \vspace{-5pt}
\end{table}

\begin{table}[t!]
    \centering
\caption{\looseness=-1 \textbf{Performance comparison on the four numerical tasks of VSI-Bench}. The best results are shown in \textbf{bold}, and the second-best are \thickunderline{underlined}. A superscript plus sign (\textbf{\textsuperscript{+}}) indicates a performance improvement over the base Qwen2.5-VL-3B model.}
\label{tab:vsibench_numerical}
\vspace{3pt}
\small
\setlength{\tabcolsep}{4pt}
\resizebox{0.9\textwidth}{!}{
\begin{tabular}{l ccccc}
\toprule
\textbf{Method} & \textbf{Obj. Count} & \textbf{Abs. Dist} & \textbf{Obj. Size} & \textbf{Room Size} & \textbf{Avg.} \\
\midrule

GPT-4o \cite{GPT-4o} & \textbf{46.2} & 5.3 & \thickunderline{43.8} & \textbf{38.2} & \thickunderline{33.4} \\
\midrule

Qwen2.5-VL-3B (Base) \cite{qwen2.5-VL} & 21.9 & \thickunderline{18.4} & 22.4 & 27.2 & 22.5 \\
\rowcolor{gray!10}
\textbf{Exemplar2VQA (Ours)} & \thickunderline{33.6}\rlap{\textbf{\textsuperscript{+}}} & \textbf{29.7}\rlap{\textbf{\textsuperscript{+}}} & \textbf{58.8}\rlap{\textbf{\textsuperscript{+}}} & \thickunderline{27.4}\rlap{\textbf{\textsuperscript{+}}} & \textbf{37.4}\rlap{\textbf{\textsuperscript{+}}} \\

\bottomrule
\end{tabular}
}

    \vspace{-5pt}
\end{table}

\subsection{Arbitrary Control and Analysis of QA Category Distributions}
\label{sec:A.4}

\textbf{The Bottleneck of Existing Benchmarks.} 
While recent benchmarks have significantly advanced the evaluation of spatial intelligence in Multimodal Large Language Models (MLLMs), relying on human-annotated datasets inevitably introduces static and imbalanced category distributions. Due to the inherent preferences of human annotators and the varying difficulty of question formulation, conventional datasets often exhibit a heavy bias towards straightforward spatial queries (e.g., basic relative direction or object recognition). When MLLMs are fine-tuned on such skewed distributions, they tend to develop strong preference biases, resulting in degraded generalization performance when confronted with rare or out-of-distribution spatial configurations.

\textbf{Exemplar2VQA's Arbitrary Control Paradigm.} 
To overcome this critical bottleneck, Exemplar2VQA fundamentally shifts the paradigm from static dataset curation to dynamic, programmable data synthesis. Because Exemplar2VQA relies on a deterministic multi-agent coding framework interacting with scene metadata, it functions as an autonomous generation engine rather than a fixed dataset. Driven by a theoretically infinite generation capacity—bounded only by the diversity and quantity of available 3D simulated scenes (as detailed in Appendix \ref{sec:A.2} via integrations with ProcTHOR and Holodeck)—Exemplar2VQA allows researchers to programmatically define the exact synthesis ratios across any spatial query categories.

\textbf{Distribution Comparison and Visualization.} 
As illustrated in Figure \ref{fig:compare} (Top), existing leading benchmarks (such as ViewSpatial-Bench, MMSI-Bench, and VSI-Bench) are constrained by rigid QA category proportions, with certain basic spatial categories dominating the visual evaluations. In stark contrast, the bottom panel of Figure \ref{fig:compare} conceptually visualizes Exemplar2VQA's arbitrary control mechanism through an intuitive ``slider'' metaphor. Mechanistically, rather than relying on complex data augmentation or rebalancing algorithms, Exemplar2VQA achieves this flexibility through sheer scale. Because the pipeline synthesizes an overwhelmingly large and diverse pool of candidate QA pairs across thousands of simulated rooms, deriving a dataset with any exact, customized category distribution is as straightforward as performing targeted random sampling from this massive pool. This effectively amplifies the presence of scarce spatial queries, effortlessly circumventing the data imbalance bottleneck.

\textbf{Implications for MLLM Spatial Intelligence.} 
This unprecedented flexibility offers profound implications for advancing the 3D spatial reasoning capabilities of Multimodal Large Language Models (MLLMs):

\textit{1. Targeted Remediation for Spatial Hallucinations:} When diagnostic benchmarks reveal that an MLLM struggles with specific geometric concepts (e.g., exhibiting severe spatial hallucinations regarding 3D occlusion, allocentric viewpoints, or relative depth ordering), researchers are no longer bound by the prohibitive costs of manual data collection. Exemplar2VQA can instantly synthesize targeted diagnostic datasets, focused supervised fine-tuning.

\textit{2. Curriculum Learning for 3D Cognition:} Exemplar2VQA’s programmable distribution naturally facilitates curriculum learning for spatial intelligence. Model training pipelines can be dynamically structured to mimic progressive cognitive development—initiating with a distribution rich in basic instance recognition and simple egocentric relative directions, before smoothly transitioning into highly complex, allocentric 3D reasoning and multi-view geometric transformations. This progressive, data-driven approach yields a much more robust and compositionally sound spatial representation within the MLLM.

\subsection{Comprehensive Error Analysis} 
\label{Error_Analysis}
To gain deeper insights into the operational bottlenecks of the Exemplar2VQA pipeline, we conduct a comprehensive failure mode analysis. We categorize the types of errors encountered during the generation process and compare their distribution between the monolithic (1 Agent) and multi-agent (4 Agents) configurations.  It is crucial to note that, as established in Section  \ref{Ablation Study}, the absolute number of failures drops significantly under the multi-agent framework; therefore, the following analysis examines the proportion of remaining failure modes.

\begin{figure}
    \centering
    \includegraphics[width=\linewidth]{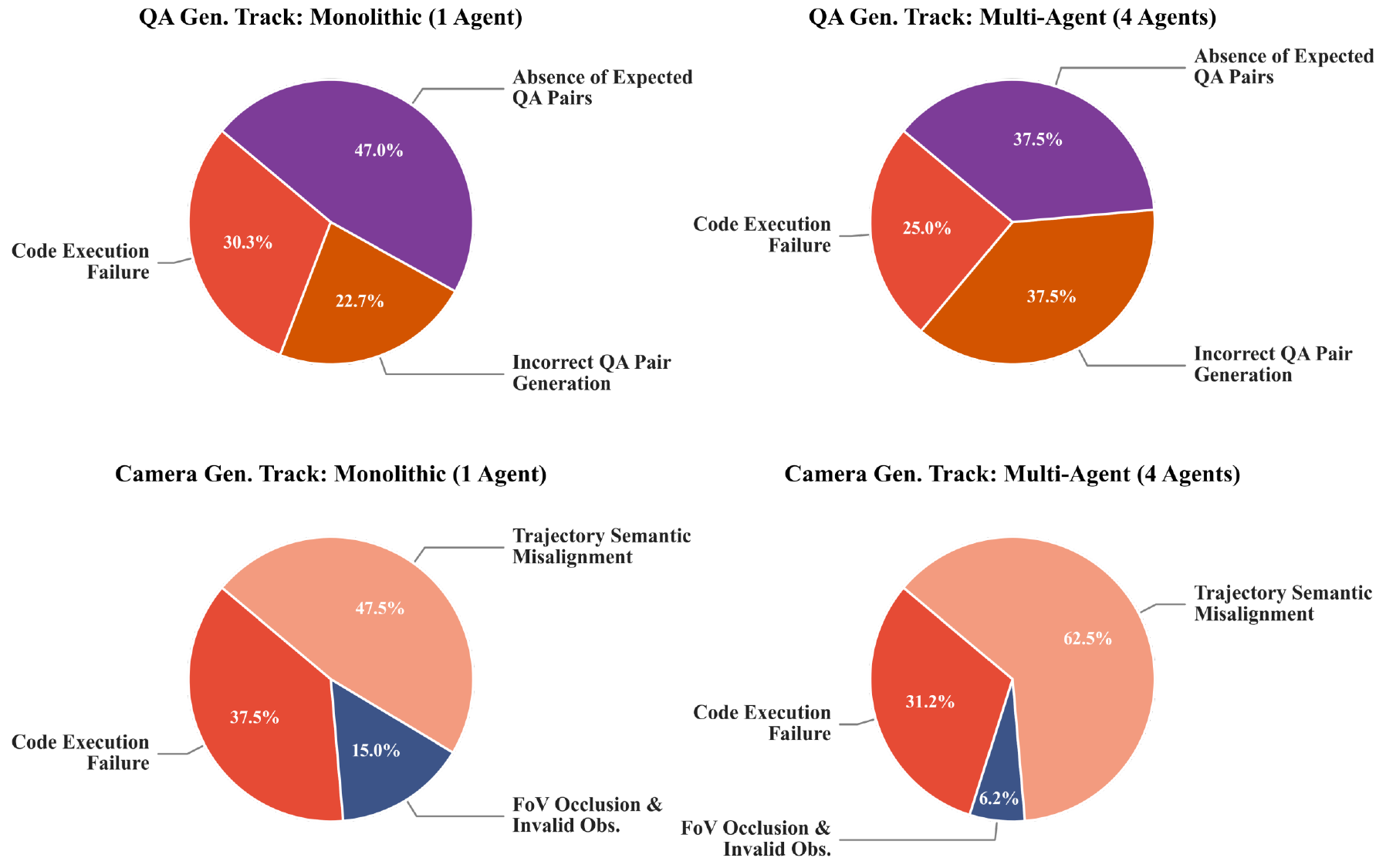}
    \caption{Distribution of error types in the Exemplar2VQA pipeline.}
    \label{fig:error}
    \vspace{-6pt}
\end{figure}

\textbf{QA Generation Track Errors.} The QA Generation Track relies on mathematical computations over scene metadata. We classify its failures into three distinct categories: \textit{Absence of Expected QA Pairs}, where the agent fails to output any valid QA pairs---despite manual inspection confirming that the scene is indeed capable of generating the expected QA pairs; \textit{Code Execution Failure}, occurring when the generated Python script crashes due to syntax errors, infinite loops, or incorrect API invocations; and \textit{Incorrect QA Pair Generation}, where the code executes successfully but the resulting QA pairs contain logical flaws or mathematically incorrect ground-truth answers. In the monolithic setup, the dominant error is the Absence of Expected QA Pairs (47.0\%), followed closely by Code Execution Failure (30.3\%). This indicates that forcing a single LLM to simultaneously act as a semantic planner, a geometric coder, and a syntax debugger leads to severe cognitive overload. Transitioning to the multi-agent setup, the distribution shifts significantly. The proportion of Code Execution Failures decreases to 25.0\%, and Absence of Expected QA Pairs drops to 37.5\%.

\textbf{Camera Trajectory Generation Track Errors.} The Camera Trajectory Generation Track interacts directly with the AI2-THOR physics engine. We categorize its failures into three types: \textit{Trajectory Semantic Misalignment}, where the generated trajectory executes successfully but fails to capture the specific visual observations requested by the abstract instruction; \textit{Code Execution Failure}, which involves Python syntax errors or logical breakdowns in the navigation loop; and \textit{FoV Occlusion \& Invalid Observations}, where the agent successfully moves, but the captured frames are invalid due to physical constraints such as staring directly into a wall, heavy occlusion, or out-of-bounds rendering. In the monolithic setup, embodied physical constraints are a major bottleneck, with Code Execution Failures (37.5\%) and FoV Occlusion \& Invalid Observations (15.0\%) collectively accounting for over half of the errors. Conversely, the multi-agent setup demonstrates a profound ability to resolve these embodied physical errors. The closed-loop physics feedback provided to the Refiner agent drastically shrinks the proportion of FoV Occlusion down to a mere 6.2\%, and Code Execution Failures drop to 31.2\%. As the pipeline reliably solves physical collisions and syntax crashes, the remaining errors become overwhelmingly dominated by Trajectory Semantic Misalignment (62.5\%). This highlights a critical frontier for Embodied AI: while multi-agent coding with physics feedback can guarantee a safe and executable trajectory, aligning an agent's continuous spatial navigation perfectly with nuanced, abstract human semantic intent remains an inherently challenging problem.

\subsection{Detailed Efficiency and Throughput Analysis} \label{sec:Efficiency}
In Section \ref{Remarks_on_Efficiency}, we briefly discussed the efficiency of the Exemplar2VQA framework. A critical advantage of Exemplar2VQA's multi-agent coding paradigm is the structural decoupling of \textit{LLM-driven Code Synthesis} (Phase 1) from \textit{Simulator-driven Data Execution} (Phase 2). This separation ensures that the computationally expensive LLM inference acts strictly as a one-time initial overhead per spatial query template, while the subsequent large-scale dataset generation operates as a highly scalable programmatic execution.

\textbf{Phase 1: Code Synthesis.} 
This phase relies on the dual-track multi-agent system to interpret semantic templates and output bug-free Python scripts.
\begin{itemize}
    \item \textbf{Camera Trajectory Track:} It takes an average of \textbf{3 minutes} per novel spatial task to complete the entire pipeline: semantic planning, code generation, and iterative interactive debugging within the AI2-THOR physics engine (e.g., resolving wall collisions or out-of-bounds errors).
    \item \textbf{QA Generation Track:} Because this track relies on mathematical computations over scene metadata rather than embodied navigation, the agents can typically synthesize and debug the optimal geometric QA code in approximately \textbf{2 minute} per task.
\end{itemize}

\textbf{Phase 2: Data Execution.}
Once the verified Python scripts ($z'_{\mathrm{cam}}$ and $z_{\mathrm{qa}}$) are synthesized, Exemplar2VQA completely bypasses the LLMs. The data generation process scales seamlessly across varying numbers of 3D scenes.
\begin{itemize}
    \item \textbf{Camera Rendering Execution:} Executing the generated camera trajectory script to navigate, render, and save the required multi-view images takes approximately \textbf{15 seconds} per 3D scene.
    \item \textbf{QA Pair Instantiation:} To maximize the yield of generated QA pairs, the LLM-synthesized scripts frequently employ brute-force enumeration algorithms. While this can theoretically result in high polynomial complexities (e.g., up to $\mathcal{O}(M^4)$ for complex multi-object spatial relationships, where $M$ is the number of objects), the actual number of observable objects per room is typically small ($M \approx 20$). Executing the code to instantiate QA pairs across \textbf{100 distinct scenes requires only about 25 seconds} in total.
\end{itemize}

In summary, this bipartite architecture allows Exemplar2VQA to effectively leverage the intelligence of LLMs to establish a robust programmatic foundation, subsequently relying on pure CPU/GPU arithmetic to rapidly scale the spatial dataset to theoretically infinite bounds.

\subsection{Limitations}
\label{sec:limitation}

While Exemplar2VQA establishes a robust paradigm for scalable spatial QA synthesis, its current scope is predominantly optimized for static, object-centric configurations, leaving complex dynamic spatiotemporal events (e.g., continuous motion tracking) for future exploration. Furthermore, the generation framework currently relies on explicit 3D scene metadata extracted from simulators; directly ingesting raw real-world photographs to autonomously synthesize QA pairs will require integrating advanced upstream models. Finally, because the pipeline relies on multi-agent code execution to bypass spatial hallucinations, its absolute success rate remains inherently bounded by the coding capabilities of the foundational LLM, particularly when navigating highly convoluted algorithmic edge cases or physics engine glitches.

\subsection{Qualitative Examples of Synthesized Spatial QA Pairs} 
\label{sec:qa_example}

To further illustrate the versatility and generative fidelity of the Exemplar2VQA framework, this section provides a comprehensive gallery of qualitative examples. The following figures showcase a diverse array of 3D simulated environments alongside their corresponding, programmatically synthesized spatial QA pairs. These visualizations highlight Exemplar2VQA's capability to accurately instantiate a wide spectrum of complex geometric queries. 

\begin{figure}
    \centering
    \includegraphics[width=\linewidth]{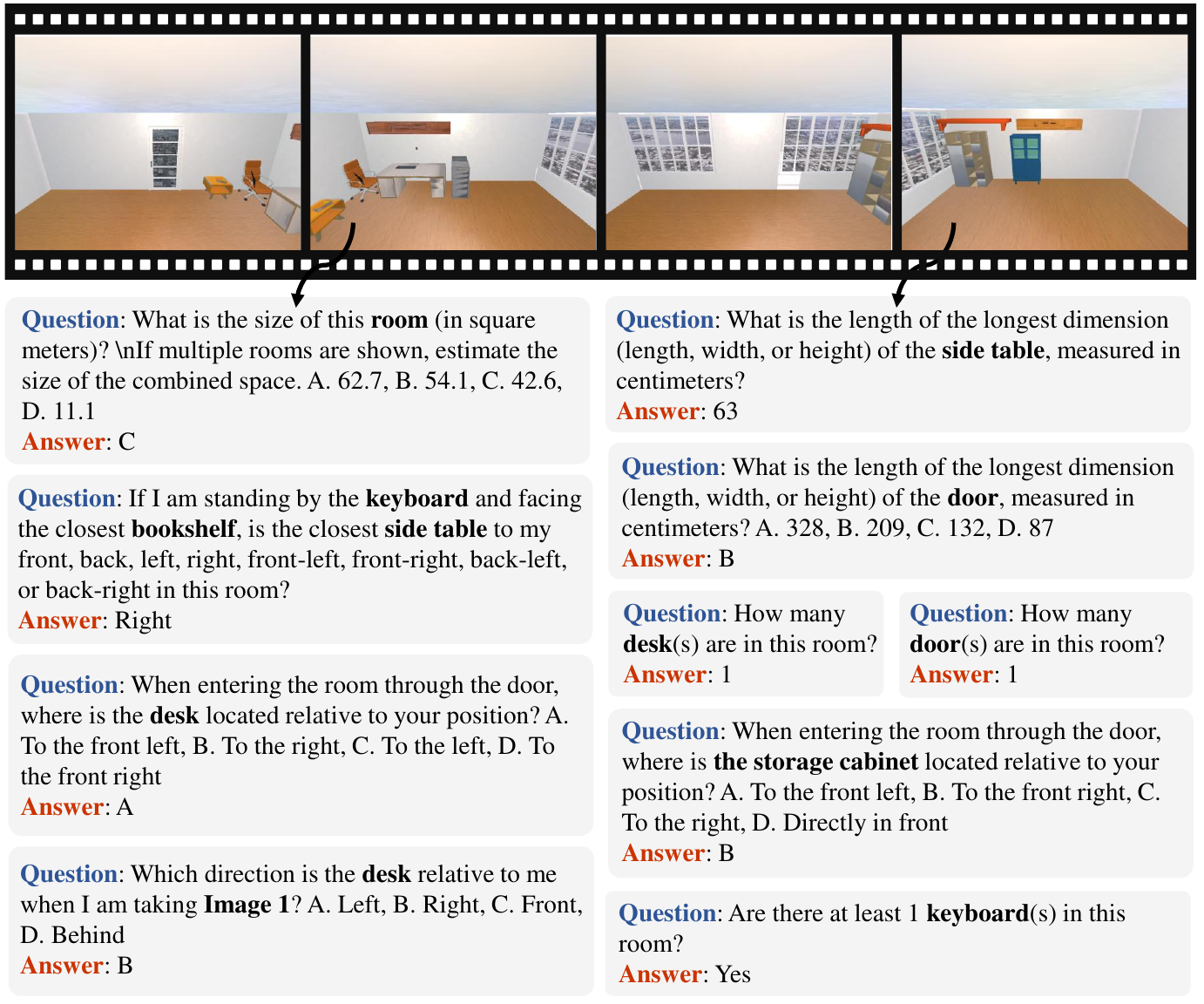}
    \caption{Additional qualitative examples of diverse spatial reasoning QA pairs autonomously synthesized across simulated environments.}
    \label{fig:keshihua_1}
    \vspace{-6pt}
\end{figure}

\begin{figure}
    \centering
    \includegraphics[width=\linewidth]{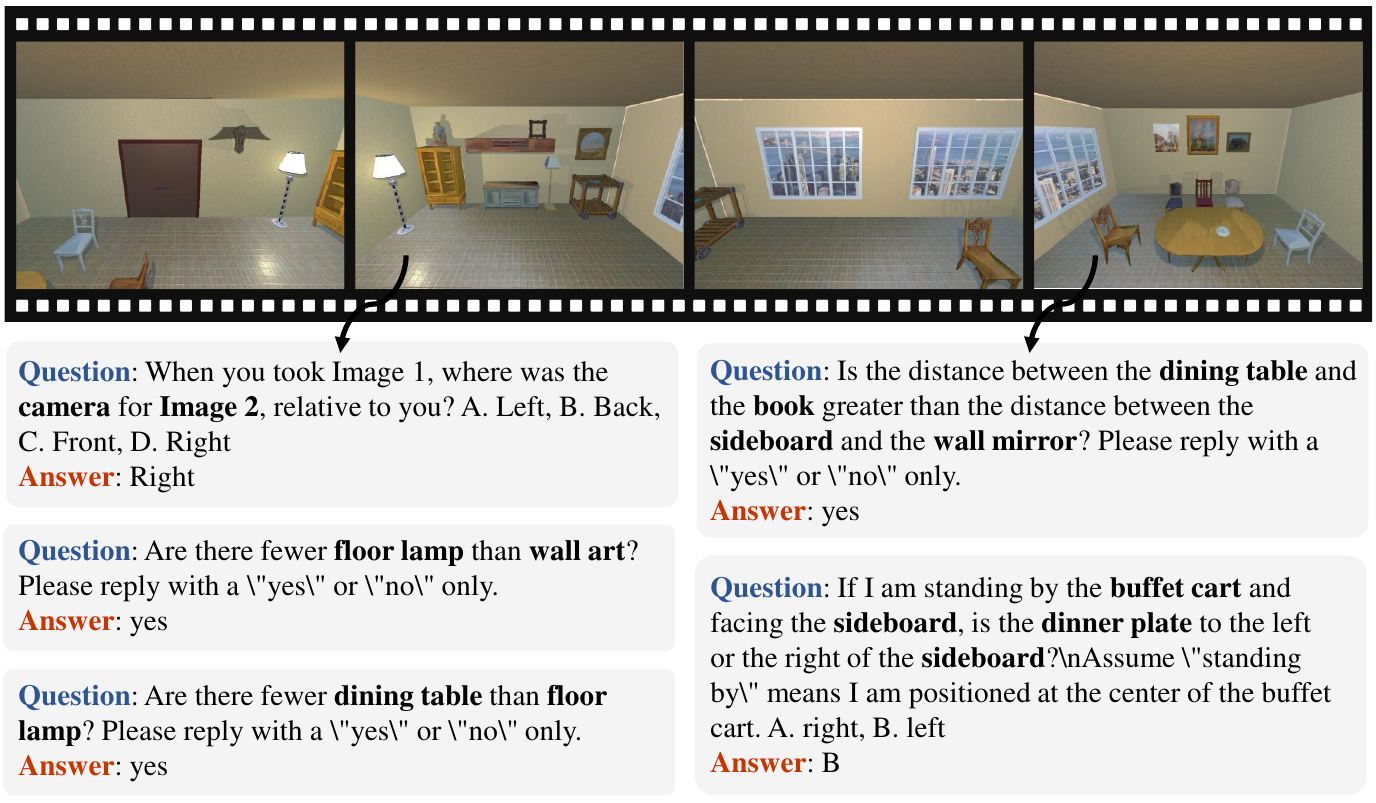}
    \caption{Additional qualitative examples of diverse spatial reasoning QA pairs autonomously synthesized across simulated environments.}
    \label{fig:keshihua_2}
    \vspace{-6pt}
\end{figure}

\begin{figure}
    \centering
    \includegraphics[width=\linewidth]{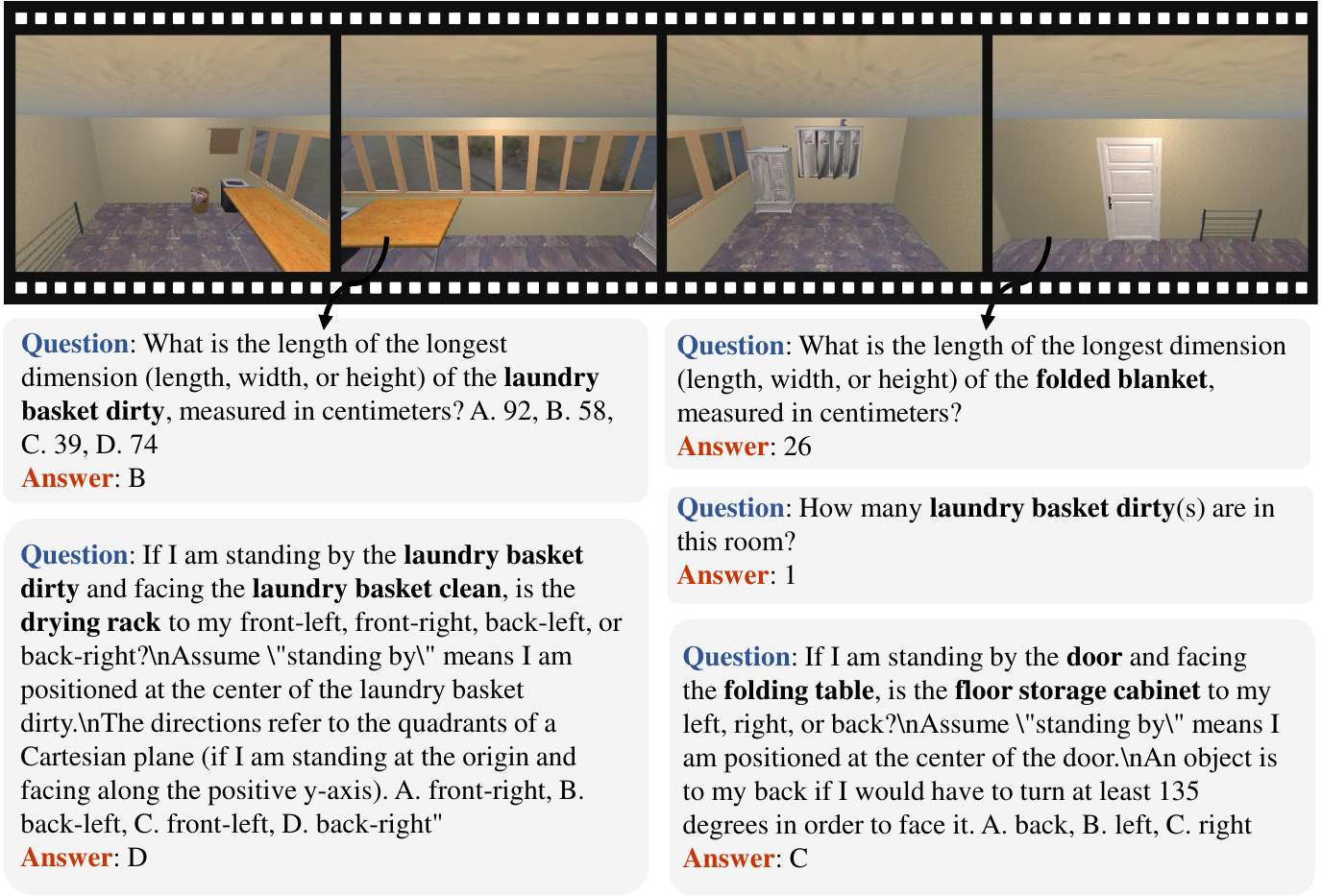}
    \caption{Additional qualitative examples of diverse spatial reasoning QA pairs autonomously synthesized across simulated environments.}
    \label{fig:keshihua_3}
    \vspace{-6pt}
\end{figure}

\begin{figure}
    \centering
    \includegraphics[width=\linewidth]{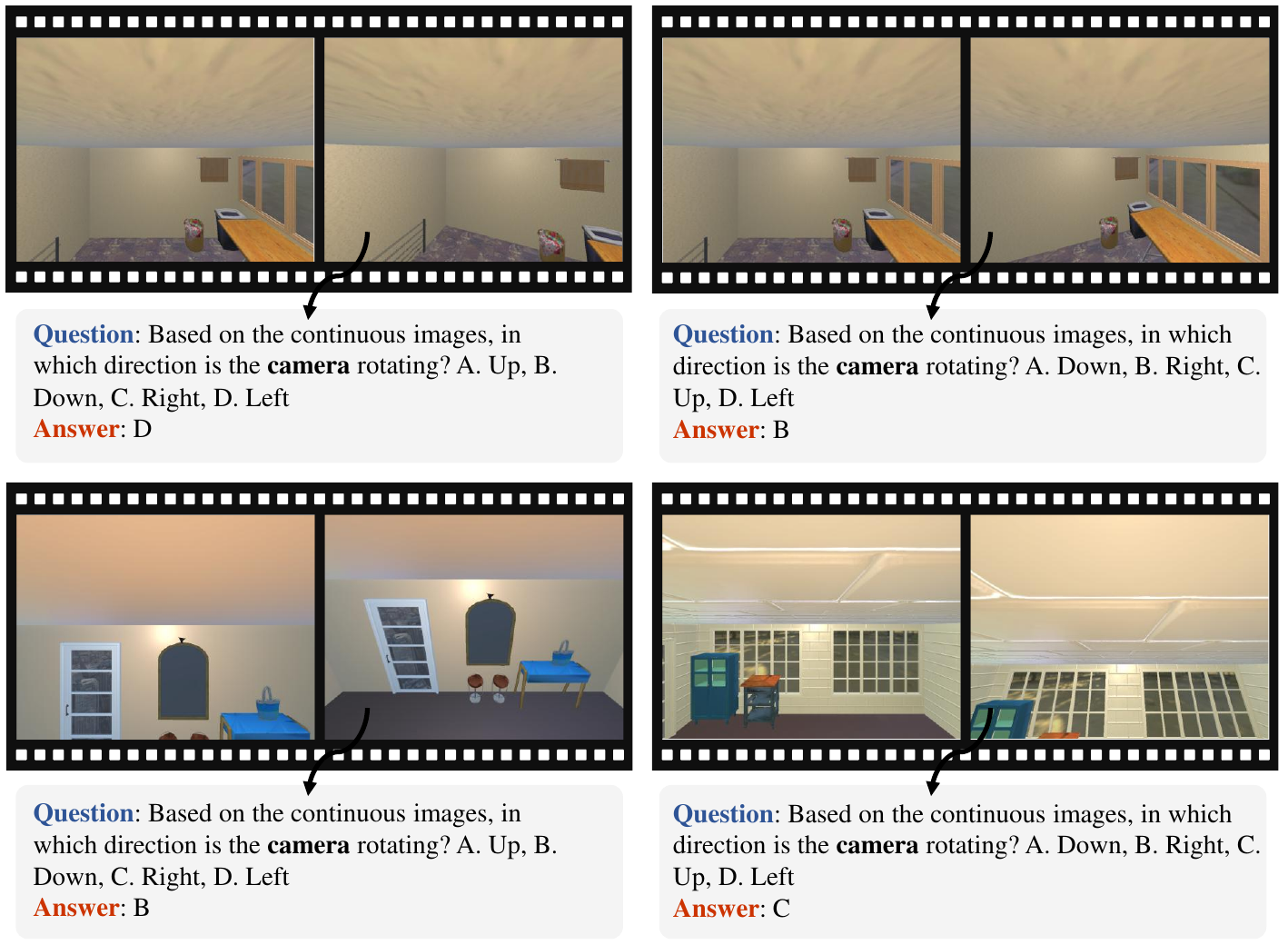}
    \caption{Additional qualitative examples of diverse spatial reasoning QA pairs autonomously synthesized across simulated environments.}
    \label{fig:keshihua_4}
    \vspace{-6pt}
\end{figure}

\begin{figure}
    \centering
    \includegraphics[width=\linewidth]{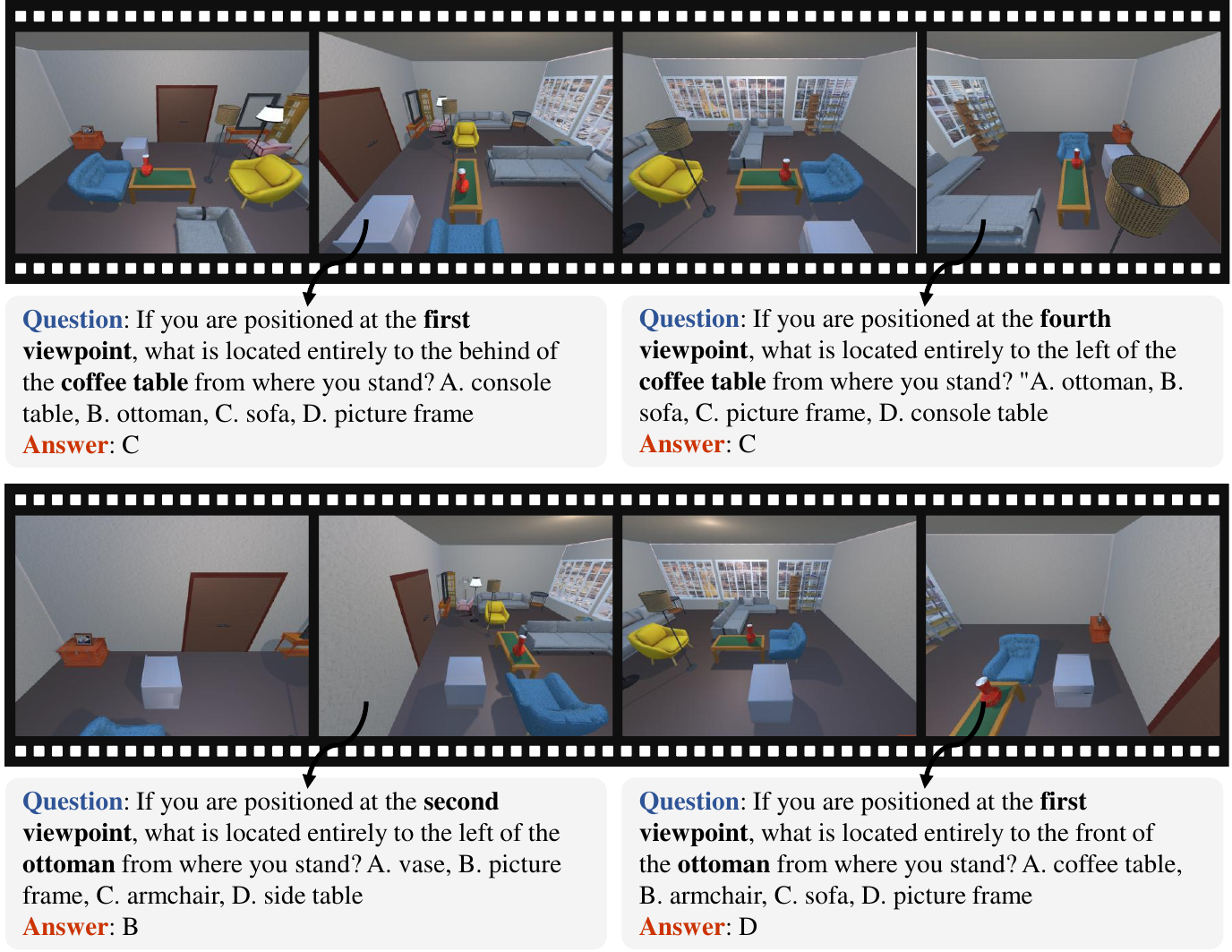}
    \caption{Additional qualitative examples of diverse spatial reasoning QA pairs autonomously synthesized across simulated environments.}
    \label{fig:keshihua_5}
    \vspace{-6pt}
\end{figure}

\begin{figure}
    \centering
    \includegraphics[width=\linewidth]{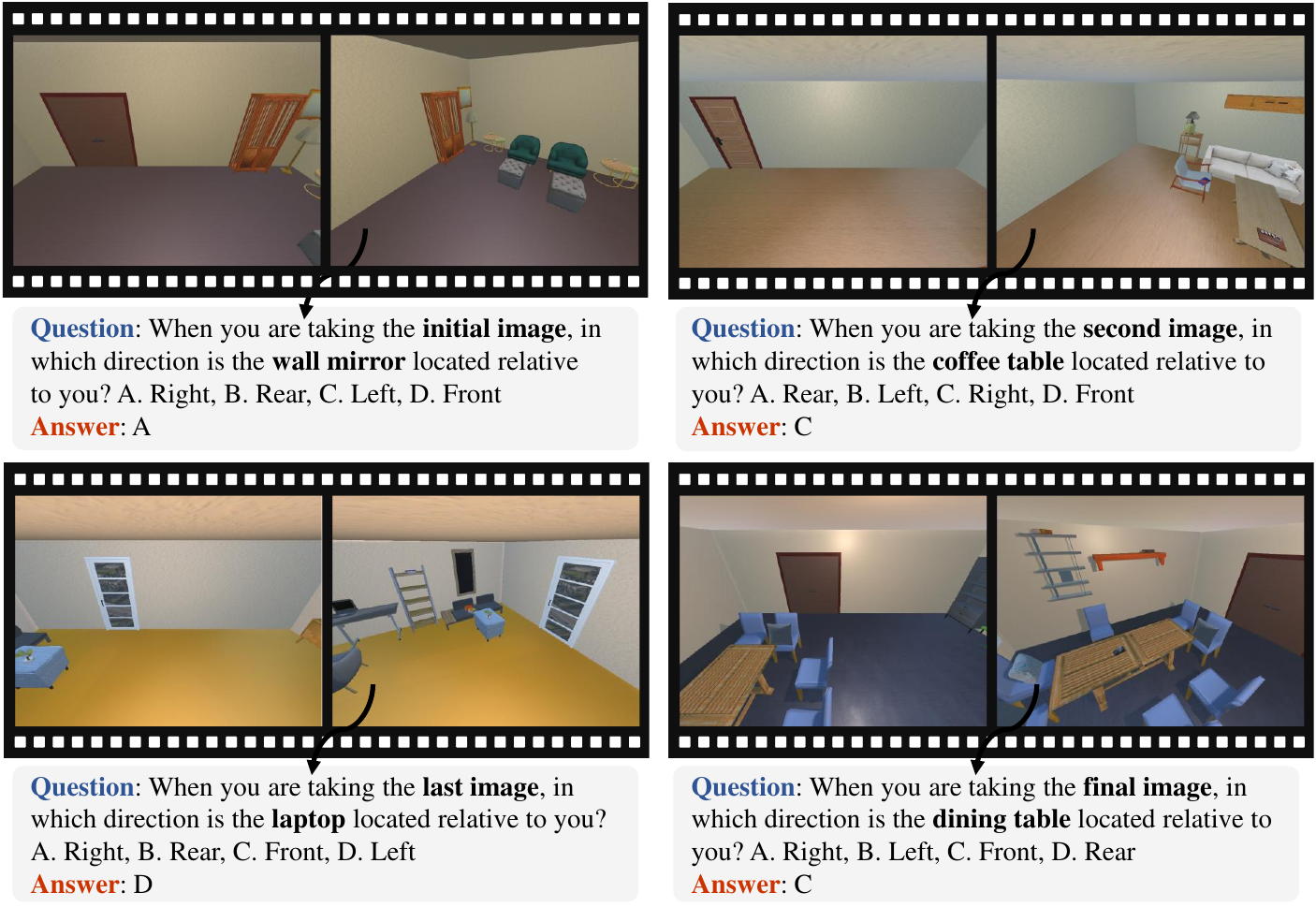}
    \caption{Additional qualitative examples of diverse spatial reasoning QA pairs autonomously synthesized across simulated environments.}
    \label{fig:keshihua_6}
    \vspace{-6pt}
\end{figure}

\clearpage

\section*{NeurIPS Paper Checklist}

\begin{enumerate}

\item {\bf Claims}
    \item[] Question: Do the main claims made in the abstract and introduction accurately reflect the paper's contributions and scope?
    \item[] Answer: \answerYes{} 
    \item[] Justification: The claims made in the abstract and introduction accurately reflect the paper's scope and contributions.

\item {\bf Limitations}
    \item[] Question: Does the paper discuss the limitations of the work performed by the authors?
    \item[] Answer: \answerYes{} 
    \item[] Justification: We discuss the limitations of this work in Appendix \ref{sec:limitation}.

\item {\bf Theory assumptions and proofs}
    \item[] Question: For each theoretical result, does the paper provide the full set of assumptions and a complete (and correct) proof?
    \item[] Answer: \answerNA{} 
    \item[] Justification: This paper focuses on the empirical design, implementation, and evaluation of the Exemplar2VQA data generation framework. It does not introduce any theoretical results, mathematical theorems, or formal proofs; therefore, this question is not applicable.

    \item {\bf Experimental result reproducibility}
    \item[] Question: Does the paper fully disclose all the information needed to reproduce the main experimental results of the paper to the extent that it affects the main claims and/or conclusions of the paper (regardless of whether the code and data are provided or not)?
    \item[] Answer: \answerYes{} 
    \item[] Justification: We fully disclose all information necessary to reproduce the main experimental results, including but not limited to the details provided in Sections \ref{Method}, \ref{Experiments}, and Appendix \ref{sec:A.1}.

\item {\bf Open access to data and code}
    \item[] Question: Does the paper provide open access to the data and code, with sufficient instructions to faithfully reproduce the main experimental results, as described in supplemental material?
    \item[] Answer: \answerYes{} 
    \item[] Justification: The code of this work will be open-sourced upon publication of the paper.

\item {\bf Experimental setting/details}
    \item[] Question: Does the paper specify all the training and test details (e.g., data splits, hyperparameters, how they were chosen, type of optimizer) necessary to understand the results?
    \item[] Answer: \answerYes{} 
    \item[] Justification: We specify all the training and test details in Section \ref{Experiments} and Appendix \ref{sec:A.1}.

\item {\bf Experiment statistical significance}
    \item[] Question: Does the paper report error bars suitably and correctly defined or other appropriate information about the statistical significance of the experiments?
    \item[] Answer: \answerNo{} 
    \item[] Justification: The primary contribution of this work is the proposal of the Exemplar2VQA data generation pipeline. Our extensive evaluations across multiple diverse benchmarks demonstrate that models fine-tuned on Exemplar2VQA-generated data achieve substantial and consistent performance improvements over baselines (e.g., an absolute gain of +7.6\% on VSI-Bench and +8.9\% on SpaCE-10). The significant magnitude of these gains across various tasks provides strong empirical evidence of the framework's efficacy, well beyond marginal statistical fluctuations. Given this robust multi-benchmark validation, coupled with the computational resources required to repeatedly fine-tune Multimodal Large Language Models (3B and 7B), performing multiple independent trials with different random seeds to compute error bars was not conducted.

\item {\bf Experiments compute resources}
    \item[] Question: For each experiment, does the paper provide sufficient information on the computer resources (type of compute workers, memory, time of execution) needed to reproduce the experiments?
    \item[] Answer: \answerYes{} 
    \item[] Justification: Computational resources are described in detail in Section \ref{Experiments} and Appendix \ref{sec:A.1}.
    
\item {\bf Code of ethics}
    \item[] Question: Does the research conducted in the paper conform, in every respect, with the NeurIPS Code of Ethics \url{https://neurips.cc/public/EthicsGuidelines}?
    \item[] Answer: \answerYes{} 
    \item[] Justification: We have thoroughly read the NeurIPS Code of Ethics and ensured compliance in all aspects of our research.

\item {\bf Broader impacts}
    \item[] Question: Does the paper discuss both potential positive societal impacts and negative societal impacts of the work performed?
    \item[] Answer: \answerYes{} 
    \item[] Justification: We present a discussion of the broader impact and societal implications of this work in Appendix \ref{sec:A.4}.
    
\item {\bf Safeguards}
    \item[] Question: Does the paper describe safeguards that have been put in place for responsible release of data or models that have a high risk for misuse (e.g., pre-trained language models, image generators, or scraped datasets)?
    \item[] Answer: \answerNA{} 
    \item[] Justification: Not involved in misusing.

\item {\bf Licenses for existing assets}
    \item[] Question: Are the creators or original owners of assets (e.g., code, data, models), used in the paper, properly credited and are the license and terms of use explicitly mentioned and properly respected?
    \item[] Answer: \answerYes{} 
    \item[] Justification: All existing assets utilized in this research, including 3D simulators, pre-trained models, and evaluation benchmarks, are properly cited in the references.

\item {\bf New assets}
    \item[] Question: Are new assets introduced in the paper well documented and is the documentation provided alongside the assets?
    \item[] Answer: \answerNA{} 
    \item[] Justification: This paper does not introduce new assets (except for data and code, which will be open-sourced upon publication of the paper).

\item {\bf Crowdsourcing and research with human subjects}
    \item[] Question: For crowdsourcing experiments and research with human subjects, does the paper include the full text of instructions given to participants and screenshots, if applicable, as well as details about compensation (if any)? 
    \item[] Answer: \answerNA{} 
    \item[] Justification: This paper does not involve crowdsourcing experiments or research with human subjects.

\item {\bf Institutional review board (IRB) approvals or equivalent for research with human subjects}
    \item[] Question: Does the paper describe potential risks incurred by study participants, whether such risks were disclosed to the subjects, and whether Institutional Review Board (IRB) approvals (or an equivalent approval/review based on the requirements of your country or institution) were obtained?
    \item[] Answer: \answerNA{} 
    \item[] Justification: This paper does not involve crowdsourcing experiments or research with human subjects.

\item {\bf Declaration of LLM usage}
    \item[] Question: Does the paper describe the usage of LLMs if it is an important, original, or non-standard component of the core methods in this research? Note that if the LLM is used only for writing, editing, or formatting purposes and does \emph{not} impact the core methodology, scientific rigor, or originality of the research, declaration is not required.
    \item[] Answer: \answerYes{} 
    \item[] Justification: The core methodology of this research relies on a multi-agent framework powered by Large Language Models. We explicitly detail the usage of the LLM in Section \ref{Method}, specifying the deployment of the open-source Qwen3-Coder-30B-A3B-Instruct model. We thoroughly describe how it is utilized across distinct agent roles (Architect, Coder, Reviewer, Refiner) to programmatically generate code and autonomously synthesize the spatial QA datasets.

\end{enumerate}

\end{document}